%% file: main.tex
\pdfoutput=1
\documentclass[11pt]{article}

\usepackage[final]{acl}
\input{math_commands.tex}

\usepackage{times}
\usepackage{latexsym}
\usepackage{hyperref}
\usepackage{url}

\usepackage[T1]{fontenc}
\usepackage[utf8]{inputenc}

\usepackage{microtype}

\usepackage{inconsolata}

\usepackage{hyperref}
\usepackage{url}
\usepackage{booktabs}       % professional-quality tables
\usepackage{amsfonts}       % blackboard math symbols
\usepackage{nicefrac}       % compact symbols for 1/2, etc.
\usepackage{microtype}      % microtypography
\usepackage{xcolor}         % colors
\usepackage{natbib}
\usepackage{amsmath, amsthm, amssymb} 
\usepackage{enumitem}
\usepackage{graphicx}
\usepackage{subcaption}
\usepackage{booktabs}
\usepackage{comment}
\usepackage{soul} 
\usepackage{algorithm}
\usepackage[noend]{algpseudocode}
\newcommand{\update}[1]{{#1}}
\newcommand{\finalchange}[1]{{#1}}

\title{Mixture-of-Supernets: Improving Weight-Sharing Supernet Training with Architecture-Routed Mixture-of-Experts}

\author{Ganesh Jawahar$^{\mu\heartsuit}$\thanks{Some of the work was completed while Ganesh was interning at Meta.} \quad Haichuan Yang$^\infty$ \quad Yunyang Xiong$^\infty$ \quad Zechun Liu$^\infty$ \\ \textbf{Dilin Wang}$^\infty$ \quad \textbf{Fei Sun}$^\infty$ \quad \textbf{Meng Li}$^\infty$ \quad \textbf{Aasish Pappu}$^\infty$ \quad \textbf{Barlas Oguz}$^\infty$ \\ \textbf{Muhammad Abdul-Mageed}$^{\mu\diamondsuit}$ \quad
 \textbf{Laks V.S. Lakshmanan}$^\mu$ \\ \textbf{Raghuraman Krishnamoorthi}$^\infty$ \quad \textbf{Vikas Chandra}$^\infty$\\
$^\mu$University of British Columbia \quad $^\infty$Meta \quad $^\diamondsuit$MBZUAI \quad $^\heartsuit$Google DeepMind \\
\texttt{ \small ganeshjwhr@gmail.com, \{haichuan, yunyang, zechunliu, wdilin, feisun, aasish, barlaso\}@meta.com}
\\ 
\texttt{ \small meng.li@pku.edu.cn, muhammad.mageed@ubc.ca, laks@cs.ubc.ca, \{raghuraman, vchandra\}@meta.com}
}

\begin{document}
\maketitle

\input{tex/abstract.tex}
\input{tex/intro.tex}
\input{tex/supernet_intro.tex}
\input{tex/proposal.tex}

\input{tex/experiments_bert.tex}

\input{tex/experiments_mt.tex}

\input{tex/related.tex}
\input{tex/conclusion.tex}
\input{tex/limitations}

% Bibliography entries for the entire Anthology, followed by custom entries
%\bibliography{anthology,custom}
% Custom bibliography entries only
\bibliography{anthology,custom}

\appendix

\input{tex/appendix}

\end{document}

%% file: math_commands.tex
\usepackage{amsmath,amsfonts,bm}

\def\eqref#1{equation~\ref{#1}}
\def\1{\bm{1}}

\DeclareMathAlphabet{\mathsfit}{\encodingdefault}{\sfdefault}{m}{sl}
\SetMathAlphabet{\mathsfit}{bold}{\encodingdefault}{\sfdefault}{bx}{n}

%% file: tex/abstract.tex
\begin{abstract}
%\gan{supernet training - pros: quick estimate, saves compute; challenges - increase supernet capacity, specialize weights for architecture; gradient conflict; large supernet vs. scratch gap;}
% Weight-sharing supernet has become a vital component for performance estimation in the state-of-the-art (SOTA) neural architecture search (NAS) frameworks.
% Although supernet can directly generate different subnetworks without retraining, there is no guarantee for the quality of these subnetworks because of weight sharing. In NLP tasks such as machine translation and pre-trained language modeling, we observe that given the same model architecture, there is a large performance gap between supernet and training from scratch. Hence, supernet cannot be directly used and retraining is necessary after finding the optimal architectures.
%To remove this limitation, we propose a framework to enhance the supernet capability to generate different weights for different subnetworks.
Weight-sharing supernets are crucial for performance estimation in cutting-edge neural architecture search (NAS) frameworks. Despite their ability to generate diverse subnetworks without retraining, the quality of these subnetworks is not guaranteed due to weight sharing. In NLP tasks like machine translation and pre-trained language modeling, there is a significant performance gap between supernet and training from scratch for the same model architecture, necessitating retraining post optimal architecture identification.

This study introduces a solution called \textit{mixture-of-supernets}, a generalized supernet formulation leveraging mixture-of-experts (MoE) to enhance supernet model expressiveness with minimal training overhead. Unlike conventional supernets, this method employs an architecture-based routing mechanism, enabling indirect sharing of model weights among subnetworks. This customization of weights for specific architectures, learned through gradient descent, minimizes retraining time, significantly enhancing training efficiency in NLP. The proposed method attains state-of-the-art (SoTA) performance in NAS for fast machine translation models, exhibiting a superior latency-BLEU tradeoff compared to HAT, the SoTA NAS framework for machine translation. Furthermore, it excels in NAS for building memory-efficient task-agnostic BERT models, surpassing NAS-BERT and AutoDistil across various model sizes. \finalchange{The code can be found at: \url{https://github.com/UBC-NLP/MoS}.}

\end{abstract}

%% file: tex/intro.tex
\section{Introduction}
\label{sec:intro}

\begin{figure*}[t!]
    \centering
    \begin{subfigure}[t]{0.25\textwidth}
        \centering
        \includegraphics[height=0.9in, width=1.1in]{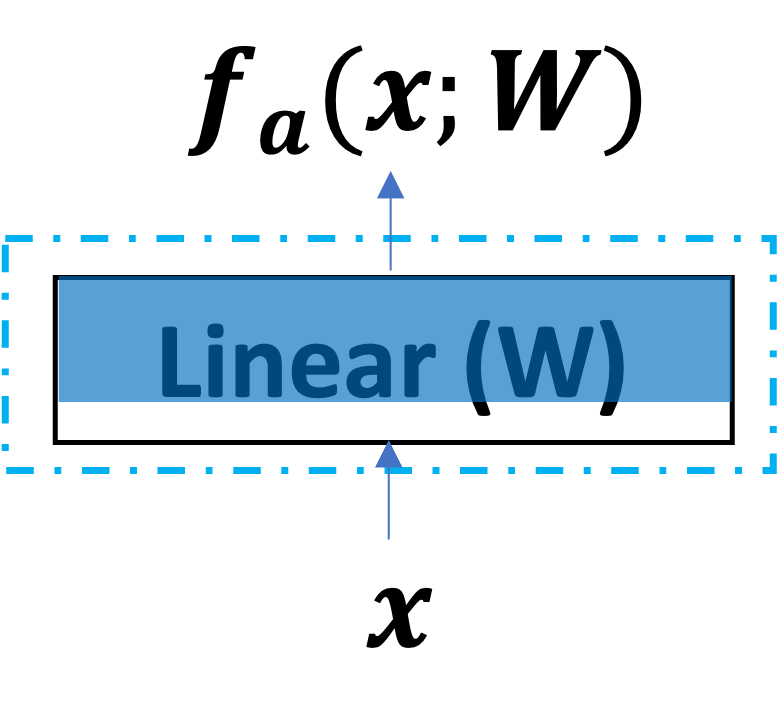}
        \caption{Standard}
    \end{subfigure}%
    ~ 
    \begin{subfigure}[t]{0.35\textwidth}
        \centering
        \includegraphics[height=1.8in, width=2.0in]{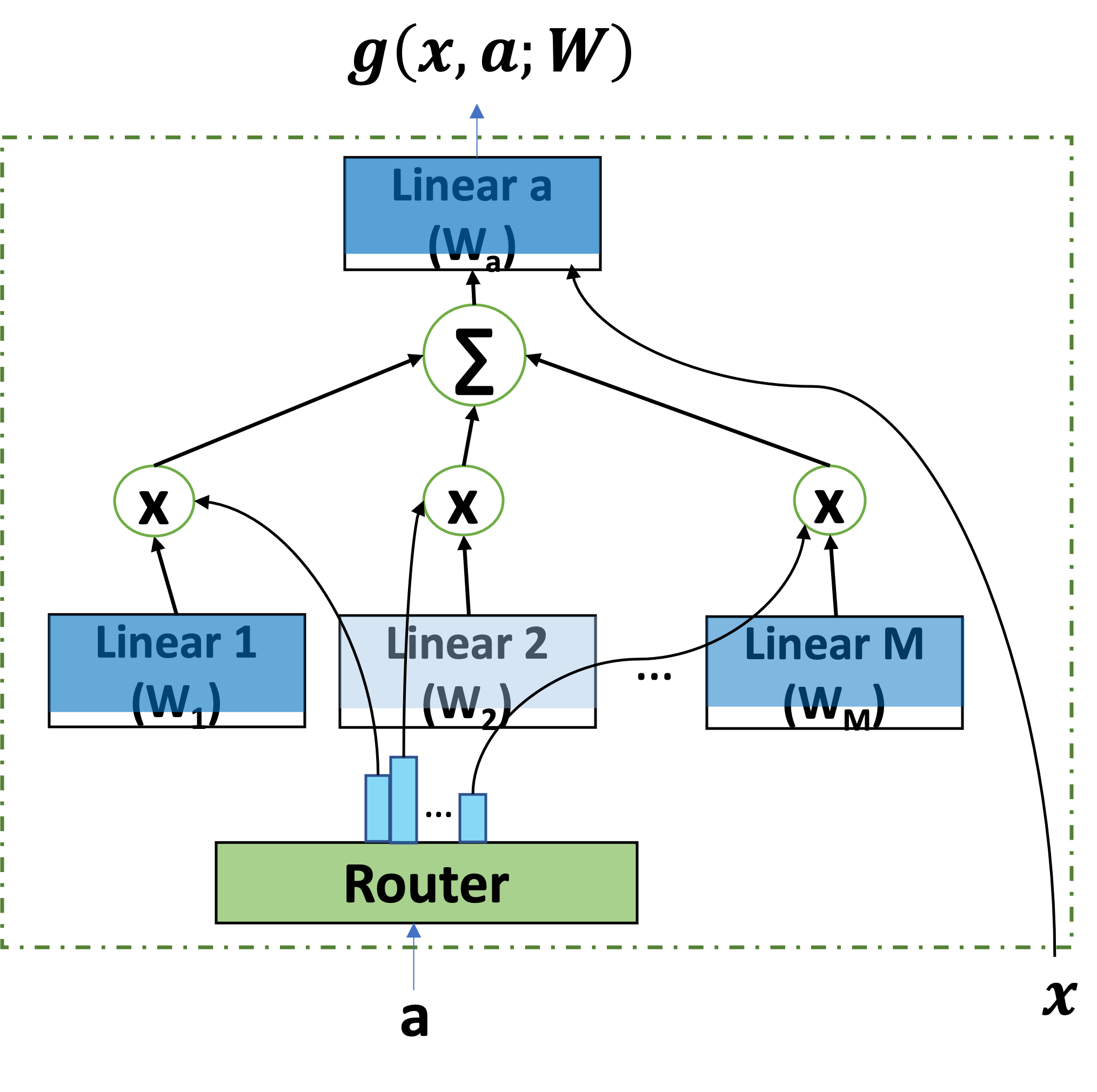}
        \caption{Layer-wise Mixture-of-Supernet}
    \end{subfigure}
    ~
    \begin{subfigure}[t]{0.35\textwidth}
        \centering
        \includegraphics[height=1.8in, width=2.0in]{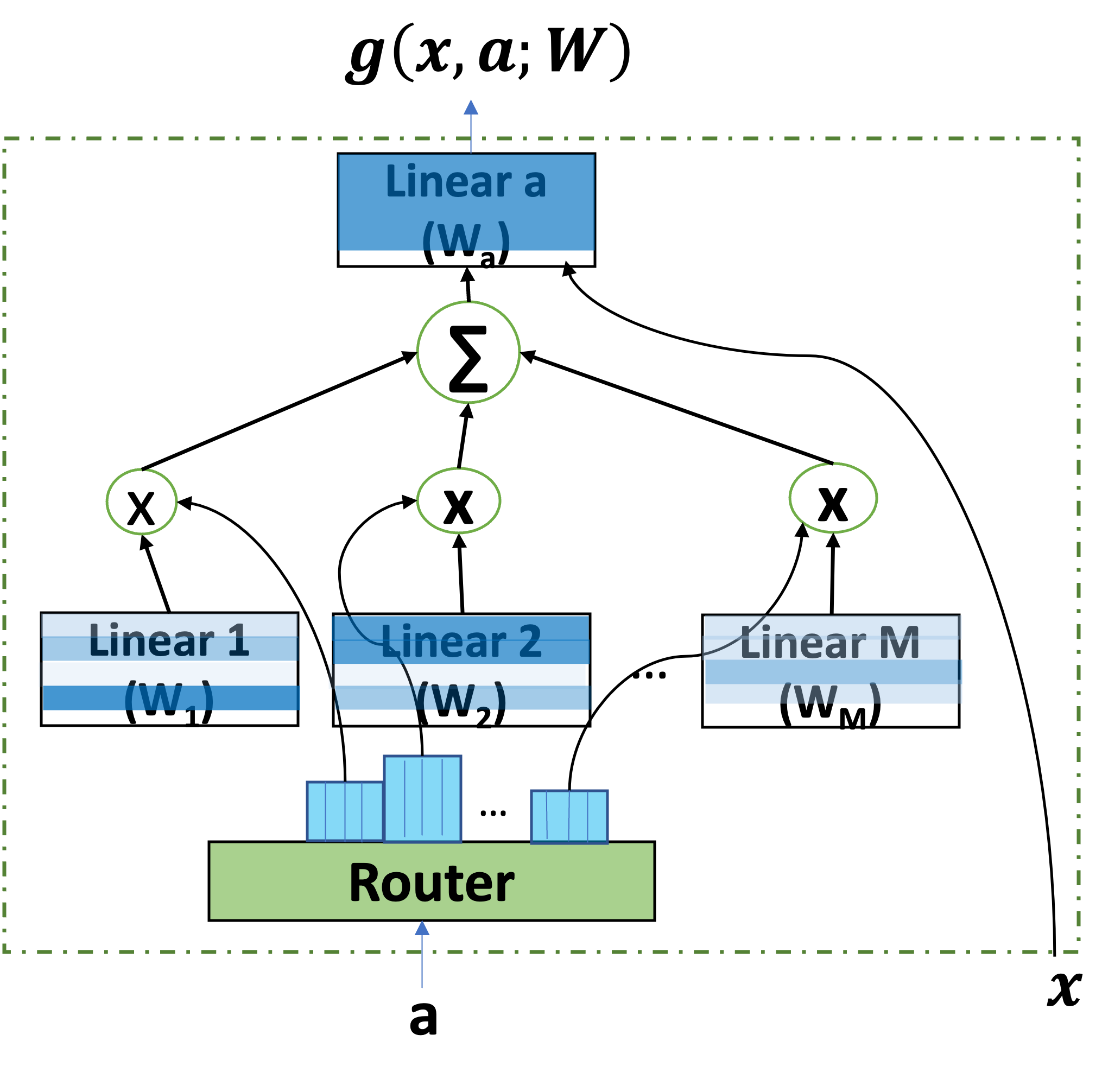}
        \caption{Neuron-wise Mixture-of-Supernet}
    \end{subfigure}
    \vspace{-0.5em}
    \caption{Choices of linear layers for supernet training. The length and the height of the `Linear' blocks correspond to the number of input and output features of the supernet respectively. The highlighted portions in blue color correspond to the architecture-specific weights extracted from the supernet. Different intensities of blue color in the `Linear' blocks of the mixture-of-supernet correspond to different alignment scores generated by the router.} % \hcnotes{words in the figure is too small and the blue color is not clear}
    \vspace{-0.5em}
    \label{fig:proposed_experts}
\end{figure*}

\begin{table*}
\scriptsize
\begin{center}
\begin{tabular}{ccccc} \toprule
\textbf{Supernet} & \textbf{Weight sharing} & \textbf{Capacity} & \textbf{Overall Time ($\downarrow$)} & \textbf{Average BLEU ($\uparrow$)} \\ \midrule %  & \textbf{Supernet GPU memory ($\downarrow$)}  % \textbf{Supernet Memory}  \\ \midrule
HAT~\citep{wang-etal-2020-hat} & Strict & Single Set & 508 hours  & 25.93 \\ % &  \textbf{10.11 GB} \\ %  & \textbf{0.98 GB} \\
Layer-wise MoS & Flexible & Multiple Set &  407 hours (20\%) & 27.21 (4.9\%) \\ % & 13.57 GB (34.2\%) \\  %& 1.4 GB \\
Neuron-wise MoS & Flexible & Multiple Set & \textbf{394 hours (22\%)} & \textbf{27.25 (5.1\%)} \\ % & 13.68 GB (35.3\%) \\ % 1.6 GB  \\
\bottomrule  
\end{tabular}
\caption{Overall time savings and average BLEU improvements of MoS supernets vs. HAT for computing pareto front (latency constraints: $100$ ms, $150$ ms, $200$ ms) for the WMT'14 En-De task. Overall time (single NVIDIA V100 hours) includes supernet training time, search time, and additional training time for the optimal architectures. Average BLEU is the average of BLEU scores of architectures in the pareto front (see Table~\ref{tab:mt_comparewithsota} for individual scores). MoS supernets yield architectures that enjoy better latency-BLEU trade-offs than HAT and have an overall GPU hours (see~\ref{sec:mt_overall_time_breakdown} for breakdown) savings of at least 20\% w.r.t. HAT. %, for at most 35\%  increase in supernet GPU memory.
% \textcolor{blue}{TODO: I am a little worried about table 1 that ppl may misunderstand that the improved BLEU score is because we evaluate a larger model. I suggest to just present the model size/memory of subnetworks, e.g., we can show the BLEU scores and model sizes of 3 models. We can move the supernet size numbers into experiment section.}
\vspace{-1.5em}
}
\label{tab:mt_search_time_savings}
\end{center}
\end{table*}

%\gan{supernet training with NAS - pros: quick estimate, saves compute; automatically design arch, withstand changes in constraints/hardware}
%Neural architecture search (NAS) can automatically design architectures that achieve high quality on the natural language processing (NLP) task, while satisfying user-defined efficiency (e.g., latency, memory) constraints~\citep{wang-etal-2020-hat,nasbert,autodistill}.
%Most straightforward way of NAS is treating it as the black-box optimization~\citep{zoph2018learning, pham2018efficient}. However, to get the architecture with the best accuracy, different model architectures need to be repeatedly trained and evaluated, which makes it impractical unless the dataset is very small. To overcome this issue, weight sharing is applied between different model architectures~\citep{pham2018efficient}. In this case, \emph{supernet} is constructed as the largest model in the search space, and each architecture is a subnetwork of it. 
Neural architecture search (NAS) automates the design of high-quality architectures for natural language processing (NLP) tasks while meeting specified efficiency constraints~\citep{wang-etal-2020-hat,nasbert,autodistill}. NAS is commonly treated as a black-box optimization~\citep{zoph2018learning, pham2018efficient}, but obtaining the best accuracy requires repetitive training and evaluation, which is impractical for large datasets. To address this, weight sharing is applied via a \emph{supernet}, where subnetworks represent different model architectures~\citep{pham2018efficient}. 
% Furthermore, recent works~\citep{onceforall, bignas} show that with good training strategies, the subnetworks can be directly used for image classification with high performance (e.g., accuracy comparable to training the same architectures from scratch).
% However, it is more challenging to apply supernet in NLP tasks. In fact, we observed that directly using the subnetworks for NLP tasks can have a large performance gap. This is consistent with the recent NAS works~\citep{wang-etal-2020-hat, nasbert} on NLP, which retrain or finetune the architectures after using supernet to find the  architecture candidates. This raises two issues: 1) it is unknown whether the selected architectures are optimal given the existence of this performance gap; 2) repeated training is still needed if we want to get the final accuracy of the Pareto front, i.e., the best models for different efficiency (e.g., model size or inference latency) budgets. In this work, we focus on improving the weight-sharing mechanism among subnetworks to minimize the performance gap.

Recent studies demonstrate successful direct use of subnetworks for image classification with performance comparable to training from scratch~\citep{onceforall, bignas}. However, applying this supernet approach to NLP tasks is more challenging, revealing a significant performance gap when using subnetworks directly. This aligns with recent NAS works in NLP~\citep{wang-etal-2020-hat, nasbert}, which address the gap by retraining or finetuning the identified architecture candidates. This situation introduces uncertainties about the optimality of selected architectures and requires repeated training for obtaining final accuracy on the Pareto front, i.e., the best models for different efficiency (e.g., model size or inference latency) budgets. This work aims to enhance the weight-sharing mechanism among subnetworks to minimize the observed performance gap in NLP tasks.

The weight-sharing supernet is trained by iteratively sampling architectures from the search space and training their specific weights from the supernet. Standard weight-sharing~\citep{bignas, onceforall} involves directly extracting the first few output neurons to create a smaller subnetwork (see Figure~\ref{fig:proposed_experts} (a)), posing two challenges due to limited model capacity. First, the supernet imposes strict weight sharing among architectures, causing co-adaptation~\citep{bender2018understanding, fewshotnas} and gradient conflicts~\citep{gong2021nasvit}. For example, in standard weight-sharing, if a 5M-parameters model is a subnetwork of a 90M-parameters model, 5M weights are directly shared. However, the optimal shared weights for the 5M model may not be optimal for the 90M model, leading to significant gradient conflicts during optimization~\citep{gong2021nasvit}. Second, the supernet's overall capacity is constrained by the parameters of a single deep neural network (DNN), i.e., the largest subnetwork in the search space. However, when dealing with a potentially vast number of subnetworks (e.g., billions), relying on a single set of weights to parameterize all of them could be insufficient~\citep{fewshotnas}.

To address these challenges, we propose a Mixture-of-Supernets (MoS) framework. MoS enables architecture-specific weight extraction, allowing smaller architectures to avoid sharing some output neurons with larger ones. Additionally, it allocates large capacity without being constrained by the number of parameters in a single DNN. MoS includes two variants: \textit{layer-wise MoS}, where architecture-specific weight matrices are constructed based on weighted combinations of expert weight matrices at the level of sets of neurons, and \textit{neuron-wise MoS}, which operates at the level of individual neurons in each expert weight matrix. Our proposed NAS method proves effective in constructing efficient task-agnostic BERT models~\citep{devlin-etal-2019-bert} and machine translation (MT) models. For efficient BERT, our best supernet outperforms SuperShaper~\citep{supershaper} by 0.85 GLUE points, surpasses NAS-BERT~\citep{nasbert} and AutoDistil~\citep{autodistill} in various model sizes ($\leq50M$ parameters). Compared to HAT~\citep{wang-etal-2020-hat}, our top supernet reduces the supernet vs. standalone model gap by 26.5\%, provides a superior pareto front for latency-BLEU tradeoff ($100$ to $200$ ms), and decreases the steps needed to close the gap by 39.8\%. A summary in the Table~\ref{tab:mt_search_time_savings} illustrates the time savings and BLEU improvements of MoS supernets for the WMT'14 En-De task.

We summarize our key contributions:
\begin{enumerate}
% \textbf{(1)} 
% We propose a formulation which can generalize weight sharing methods, including direct weight sharing (e.g., once-for-all network~\citep{onceforall}, BigNAS~\citep{bignas}) and flexible weight sharing (e.g., few-shot NAS~\citep{fewshotnas}). This formulation allows us to improve supernet by enhancing the model's expressive power. 
\item We propose a formulation that generalizes weight-sharing methods, encompassing direct weight sharing (e.g., once-for-all network~\citep{onceforall}, BigNAS~\citep{bignas}) and flexible weight sharing (e.g., few-shot NAS~\citep{fewshotnas}). This formulation enhances the expressive power of the supernet.
% \textbf{(2)} 
%We adopt the idea of MoE to improve the model capability. Specifically, the model's weights are dynamically generated based on the activated subnetwork architecture. After training, this MoE can be converted into equivalent static models. This is because our supernets only depend on the subnetwork architecture, which is fixed after training. 
\item We apply the MoE concept to enhance model capability. The model's weights are dynamically generated based on the activated subnetwork architecture. Post-training, the MoE can be converted into equivalent static models as our supernets solely depend on the fixed subnetwork architecture after training.
\textbf{(3)} 
% We conduct comprehensive experiments, demonstrating that our supernets achieve the SOTA NAS results on %both Pretrained Language Model (task-agnostic BERT) and MT tasks. 
% building efficient task-agnostic BERT and MT models.
\item  Our experiments show that our supernets achieve SoTA NAS results in building efficient task-agnostic BERT and MT models.
\end{enumerate}

%% file: tex/supernet_intro.tex
\section{Supernet - Fundamentals}
\label{sec:supernet_intro}

%\gan{Supernet - high level goal, training objective}
%The goal of training a supernet is to provide a fast performance estimate of an architecture from a given search space. 
%\update{A supernet is a model that employs weight sharing to parameterize weights for millions of architectures~\cite{wang-etal-2020-hat}. Supernet can provide quick performance predictions for various architectures, which reduces the search cost for NAS significantly.} The training objective of the supernet can be formalized as follows. 
\update{A supernet, utilizing weight sharing, parameterizes weights for millions of architectures, offering rapid performance predictions and significantly reducing NAS search costs. The training objective can be formalized as follows.}
Let $\mathcal{X}_{tr}$ denote the training data distribution. Let $x$, $y$ denote the training sample and label respectively, i.e., $x, y \sim \mathcal{X}_{tr}$. Let $a_{rand}$ denote an architecture uniformly sampled from the search space $\mathcal{A}$. Let $f_{a}$ denote the subnetwork with architecture $a$, and $f$ be parameterized by the supernet model weights $W$. Then, the training objective of the supernet can be given by,
\begin{equation}
    \min_W \mathbb{E}_{x,y \sim \mathcal{X}_{tr}} \mathbb{E}_{a_{rand} \sim \mathcal{A}} [ \mathcal{L}(f_{a_{rand}}(x;W),y)].
\label{eq:supernetobj}
\end{equation}
%\gan{Existing supernet improvement techniques: (i) SPOS - single path optimization, (ii) Sandwich sampling}
%The above formulation is known as single path one-shot (SPOS) optimization~\citep{spos_guo20} of supernet training. Sandwich training~\citep{bignas} is another popular technique for training a supernet, where the largest architecture ($a_{big}$), the smallest architecture ($a_{small}$), and the architecture ($a_{rand}$) uniformly sampled from the search space are jointly optimized. The training objective of the supernet then becomes:
The mentioned formulation is termed single path one-shot (SPOS) optimization~\citep{spos_guo20} for supernet training. Another popular technique is sandwich training~\citep{bignas}, where the largest ($a_{big}$), smallest ($a_{small}$), and uniformly sampled architectures ($a_{rand}$) from the search space are jointly optimized. 
% The supernet's training objective is then:
\begin{comment}
\begin{multline}
    \min_W \mathbb{E}_{x,y \sim \mathcal{X}_{tr}}[\mathbb{E}_{{a_{rand}} \sim \mathcal{A}} [ \mathcal{L}(f_{a_{rand}}(x;W),y)] \\ + \mathcal{L}(f_{a_{big}}(x;W),y) + \mathcal{L}(f_{a_{small}}(x;W),y)].
\end{multline}
\end{comment}

% Inplace KD

%% file: tex/proposal.tex
\section{Mixture-of-Supernets}
\label{sec:proposed_methods}

%\gan{Issues with existing supernets: limited model capacity problem; provide an example model function; provide an example for top rows extraction; provide theory for any two architectures, one arch is subset issue}
Existing supernets typically have limited model capacity to extract architecture-specific weights. For simplicity, assume the model function $f_a(x;W)$ is a fully connected layer (output $o = Wx$, omitting bias term for brevity), where $x \in n_{in}\times1$, $W \in n_{out}\times n_{in}$, and $o \in n_{out}\times1$. $n_{in}$ and $n_{out}$ correspond to the number of input and output features respectively.
%Assume the number of output features ($n_{out}$) is the only elastic hyperparameter in the search space~\cite{yu2019universally}.
Then, the weights ($W_a \in n_{{out}_a}\times n_{in}$) specific to architecture $a$ with $n_{{out}_a}$ output features are typically extracted by taking the first $n_{{out}_a}$ rows\footnote{We assume the number of input features remains constant. If it changes, only the initial columns of $W_a$ are extracted.} (as shown in Figure~\ref{fig:proposed_experts} (a)) from the supernet weight $W$. Assume one samples two architectures ($a$ and $b$) from the search space with the number of output features $n_{{out}_a}$ and $n_{{out}_b}$ respectively. Then, the weights corresponding to the architecture with the smallest number of output features will be a subset of those of the other architecture, sharing the first $|n_{{out}_a} - n_{{out}_b}|$ output features exactly. 
%\gan{Issues with existing supernets: provide an example for subset issue, intro. issue about capacity limited by hyperparameter. provide an example for this issue. Mention why common capacity increment techniques do not work.} 
%Such a weight extraction technique enforces a strict notion of weight sharing between architectures, regardless of the global architecture information (e.g., different number of features for all the other layers) of these architectures. For instance, architectures $a$ and $b$ can have widely different model capacities (e.g., $5M$ vs $90M$ number of architecture-specific parameters). The smaller architecture (e.g., $5M$) has to share all its weights with the other architecture (e.g., $90M$) and the supernet (as modeled by $f_a(x;W)$) cannot allocate any weights that are specific to the smaller architecture only. 
This weight extraction technique enforces strict weight sharing between architectures, irrespective of their global architecture information (e.g., different features in other layers). For example, architectures $a$ and $b$ may have vastly different capacities (e.g., $5M$ vs $90M$ parameters). The smaller architecture (e.g., $5M$) must share all weights with the larger one (e.g., $90M$), and the supernet (modeled by $f_a(x;W)$) cannot allocate weights specific to the smaller architecture.
% Another problem with $f_a(x;W)$ is that the overall capacity of the supernet is bounded by the number of parameters in the largest subnetwork (i.e. $W$) from the search space. However, the supernet weights $W$ need to parameterize a large amount of different subnetworks in the search space. This is a fundamental limitation of the standard weight sharing mechanism. 
Another issue with $f_a(x;W)$ is that the supernet's overall capacity is constrained by the parameters in the largest subnetwork ($W$) in the search space. Yet, these supernet weights $W$ must parameterize numerous diverse subnetworks. This represents a fundamental limitation of the standard weight-sharing mechanism.
% \update{Section~\ref{sec:gen_model_fn} proposes a reformulation to address this limitation, which is instantiated using two methods (Layer-wise MoS, Section~\ref{sec:layer_wise_mos}, Neuron-wise MoS, Section~\ref{sec:neuron_wise_mos}) and can be dropped into Transformers (see Section~\ref{sec:add_to_trans}). We conclude this section with a comparison of Mixture-of-Supernets framework to AutoMoE framework~\cite{jawahar-etal-2023-automoe} in Section~\ref{sec:c4_comparison_to_automoe}.}
% architecture-specific hyperparameters (e.g., number of output features).
\update{Section~\ref{sec:gen_model_fn} proposes a reformulation to overcome this limitation, implemented through two methods (Layer-wise MoS, Section~\ref{sec:layer_wise_mos}, Neuron-wise MoS, Section~\ref{sec:neuron_wise_mos}), suitable for integration into Transformers (see Section~\ref{sec:add_to_trans}).}

%If $n_{{in}_a}=512$ and $n_{{out}_a}=1024$, the capacity of architecture $a$ will be $1024\times512=524288$.
%Popular techniques to increase capacity of a model function such as adding hidden layers, hidden neurons are not applicable, as we cannot change the model architecture of mapping $x$ to $f_a(x;W)$.

\subsection{Generalized Model Function}
%\gan{Generalized model function - introduce two properties. New supernet objective function.}
\label{sec:gen_model_fn}
We can reformulate the function $f_a(x;W)$ to  a generalized form $g(x,a;E)$, which takes 2 inputs: the input data $x$, and the activated architecture $a$. $E$ includes the learnable parameters of $g$.
%In this work, we introduce a generalized model function $g(x,a;W)$ that has two desirable properties: (i) the function can perform architecture-specific weight extraction from the supernet (\textit{customization property}) and (ii) the expressiveness of the function in terms of the number of architecture-specific parameters can be significantly increased, not limited by the hyperparameters of the architecture (\textit{expressiveness property}).
Then, the training objective of the proposed supernet becomes,  
\begin{equation}
    \min_{{E}} \mathbb{E}_{x,y \sim \mathcal{X}_{tr}} \mathbb{E}_{a_{rand} \sim \mathcal{A}} [ \mathcal{L}(g(x,a_{rand};E),y)].
    \label{eq:propobj}
\end{equation}

For the standard weight sharing mechanism mentioned above, $E=W$ and function $g$ just uses $a$ to perform the ``trimming'' operation on the weight matrix $W$, and forwards the subnetwork. %To further minimize the objective~\eqref{eq:propobj}, one feasible way is improving the capacity of the model function $g$. However, common ways such as adding hidden layers or hidden neurons are not applicable here, as we cannot change the final subnetwork architecture of mapping $x$ to $f_a(x;W)$.
 %In this work, we propose to use the idea of Mixture-of-Experts (MoE)~\citep{switch} % JacobsJordanNowlanEtAl91
%to improve the capacity of $g$. Specifically, we dynamically generate the weights $W_{a}$ according to specific architecture $a$ by routing to certain weights matrices from a set of expert weights. We call this architecture-routed MoE based supernet  \textit{Mixture-of-Supernets (MoS)}, and design two routing mechanisms for function $g(x,a;E)$. %, which we describe next.
To further minimize objective~\eqref{eq:propobj}, enhancing the capacity of the model function $g$ is a potential approach. However, conventional methods like adding hidden layers or neurons are impractical here since the final subnetwork architecture of mapping $x$ to $f_a(x;W)$ cannot be altered.
This work introduces the concept of Mixture-of-Experts (MoE)~\citep{switch} to enhance the capacity of $g$. Specifically, we dynamically generate weights $W_{a}$ for a specific architecture $a$ by routing to certain weight matrices from a set of expert weights. We term this architecture-routed MoE-based supernet as \textit{Mixture-of-Supernets (MoS)} and design two routing mechanisms for function $g(x,a;E)$.
{Due to lack of space, the detailed algorithm for supernet training and search is shown in~\ref{sec:detailed_algorithm}.}

\subsection{Layer-wise MoS}
\label{sec:layer_wise_mos}
%\gan{layer-wise MoS - explain the theory. provide example on how it solves  previously discussed issues.}
Assume there are $m$ \update{(number of experts)} unique weight matrices ($\{{E^i \in \mathcal{R}^{n_{{out}_{big}} \times n_{{in}_{big}} }}\}^m_{i=1}$, or expert weights), which are learnable parameters. For simplicity, we only use a single linear layer as the example. For an architecture $a$ with $n_{out_a}$ output features, we propose the layer-wise MoS that computes the weights specific to the architecture $a$ (i.e. $W_a \in \mathcal{R}^{n_{out_a} \times n_{in}} $) by performing a weighted combination of expert weights,  $W_a = \sum_i \alpha^i_a E^i_a$.
%as follows: 
%\begin{equation}
%    W_a = \sum_i \alpha^i_a E^i_a.
%\end{equation}
Here, $E^i_a \in \mathcal{R}^{n_{out_a} \times n_{in}}$ corresponds to the standard top rows extraction from the $i^{th}$ expert weights.
The alignment vector ($\alpha_a \in  [0, 1]^{m}, \sum_i \alpha^i_a=1$) captures the alignment scores of the architecture $a$ with respect to each expert (weights matrix). We encode the architecture $a$ as a numeric vector $\text{Enc}(a) \in \mathcal{R}^{n_{enc}\times 1}$ (e.g., a list of the number of output features for different layers), and apply a learnable router $r(\cdot)$ (an MLP with softmax) to produce such scores, i.e. $\alpha_a = r(\text{Enc}(a))$.
% which is obtained as follows,
% \begin{equation}
% \small
%     [\alpha^1_a,\dots, \alpha^M_a] = \text{Softmax}( R_2 ReLU(R_1 \text{Enc}(a)) ) . 
% \end{equation}
%The function $\text{Enc}(\cdot)$ computes the architecture encoding ($\in \mathcal{R}^{n_{enc}\times 1}$) of a given architecture, typically the distinctive features of the architecture (e.g., a list of architecture-specific values for variable hyperparameters in the search space).
%$R_1 \in  \mathcal{R}^{n_{rhid} \times {n_{enc}}} $ and $R_2 \in  \mathcal{R}^{M \times {n_{rhid}}} $ are the learnable parameters of the router. $ReLU(.)$ function corresponds to the non-linearity function, while $Softmax(.)$ function converts the router logits to probability distribution across experts.
Thus, the generalized model function for the linear layer (as shown in Figure~\ref{fig:proposed_experts} (b)) can be defined as (omitting bias for brevity):
\begin{equation}
     g(x,a;E) = W_a x = \sum_i r(\text{Enc}(a))^i  E^i_a x.
\end{equation}
% To summarize the theory, layer-wise MoS uses architecture information to construct architecture-specific weights from a weighted combination of expert weights. Let us now see how the layer-wise MoS method solves the existing challenges of supernet. 
% Router $r(\cdot)$ controls the degree of weight sharing (unsharing) between two architectures by modulating the alignment scores ($\alpha_a$). For example, if $m=2$ and $a$ is a subnetwork of the architecture $b$, the supernet could allocate weights that are specific to the smaller architecture $a$ only by setting $\alpha_{a} = (1, 0)$ and $\alpha_{b} = (0, 1)$. In this case, $g(x,a;E)$ only uses weights from $E_1$ and $g(x,b;E)$ only uses weights from $E_2$, so $E_1$ and $E_2$ can be updated towards the loss from architecture $a$ and $b$ without conflicts. It should be noted that few-shot NAS~\citep{zhao2021few} can be seen as a special case of our framework if the router $r$ is rule-based. In addition, $g(\cdot)$ is essentially an MoE so that it has stronger expressive power and can lead the objective~\eqref{eq:propobj} to be smaller.
The router $r(\cdot)$ governs the degree of weight sharing between two architectures through modulation of alignment scores ($\alpha_a$). For instance, if $m=2$ and $a$ is a subnetwork of architecture $b$, the supernet can allocate weights specific to the smaller architecture $a$ by setting $\alpha_{a} = (1, 0)$ and $\alpha_{b} = (0, 1)$. In this scenario, $g(x,a;E)$ exclusively utilizes weights from \finalchange{$E^1$}, and $g(x,b;E)$ uses weights from \finalchange{$E^2$}, enabling updates to \finalchange{$E^1$} and \finalchange{$E^2$} towards the loss from architectures $a$ and $b$ without conflicts. It's worth noting that few-shot NAS~\citep{fewshotnas} is a special case of our framework when the router $r$ is rule-based. Moreover, $g(\cdot)$ functions as an MoE, enhancing expressive power and reducing the objective~\eqref{eq:propobj}.
%For the linear layer case, the overall capacity of the architecture allocated by $g(x,a;E)$ can be given by $n_{out_a} \times n_{in_a} \times m$, which is not limited only by the architecture-specific hyperparameters ($n_{out_a}$ and $n_{in_a}$). $m$ is an architecture-agnostic hyperparameter, which can be used to add more capacity to the architecture.
%For example, if  $n_{out_a}=1024$ and $n_{in_a}=512$, varying $m$ from 2 to 5 sets architecture capacity from $1m$ to $2m$ parameters.
% After the supernet training completes, given an architecture $a$, the score $\alpha_a = r(\text{Enc}(a))$ can be generated offline. Expert weights are collapsed and the resulting number of parameters for the architecture $a$ becomes $n_{out_a} \times n_{in_a}$. Layer-wise MoS induces low degree of weight sharing between differently sized architectures shown by higher Jensen-Shannon distance between their alignment probability vectors compared to that of similarly sized architectures. See~\ref{sec:layer_wise_sharing} for more details.
Once supernet training is done, for a given architecture $a$, the score $\alpha_a = r(\text{Enc}(a))$ can be generated offline. Expert weights collapse, reducing the number of parameters for architecture $a$ to $n_{out_a} \times n_{in_a}$. Layer-wise MoS results in a lower degree of weight sharing between differently sized architectures, as evidenced by a higher Jensen-Shannon distance between their alignment probability vectors compared to similarly sized architectures. Refer to~\ref{sec:layer_wise_sharing} for more details.

\subsection{Neuron-wise MoS}
\label{sec:neuron_wise_mos}
%The layer-wise MoS follows a conventional MoE setup, i.e., each expert is a linear layer/module. The router decides to use which experts combination to forward the input $x$ to, depending on $a$. In this case, the degree of freedom of weights generation is $m$, and the number of parameters grows by ${{m}}\times |W|$, where $|W|$ denotes the number of parameters in the standard supernet. Thus we need $m$ to be large enough to keep a good flexibility for the subnetwork weights generation, but this will also introduce too many parameters into the supernet and make the layer-wise MoS hard to train.
%This motivates us to use a smaller granularity of weights to represent each expert. Specifically, we use neurons in DNN as  experts. In terms of the weight matrix, neuron-wise MoS uses one row of matrix to represent an individual expert. In contrast, layer-wise MoS uses an entire weight matrix.

Layer-wise MoS employs a standard MoE setup, where each expert is a linear layer/module. The router determines the combination of experts to use for forwarding the input $x$ based on $a$. In this setup, the degree of freedom for weight generation is $m$, and the parameter count grows by ${{m}}\times |W|$, with $|W|$ being the parameters in the standard supernet. Therefore, a sufficiently large $m$ is needed for flexibility in subnetwork weight generation, but it also introduces too many parameters into the supernet, making layer-wise MoS challenging to train.
To address this, we opt for a smaller granularity of weights to represent each expert, using neurons in DNN as experts. In terms of the weight matrix, neuron-wise MoS represents an individual expert with one row, whereas layer-wise MoS uses an entire weight matrix.
For neuron-wise MoS, the router output $\beta_a = r(\cdot) \in [0, 1]^{n_{out_{big}} \times m}$ for each layer, and the sum of each row in $\beta_a$ is 1. Similar to layer-wise MoS, we use an MLP to produce the $n_{out_{big}} \times m$ matrix and apply softmax on each row. We formulate the function $g(x,a;E)$ for neuron-wise MoS as
\begin{equation}
    W_a = \sum_i \text{diag}(\beta^i_a) E^i_a,
\end{equation}
where $\text{diag}(\beta)$ constructs a $n_{out_{big}} \times n_{out_{big}}$ diagonal matrix by putting $\beta$ on the diagonal, and $\beta_a^i$ is the $i$-th column of $\beta_a$. $E^i$ is still an $n_{out_{big}} \times n_{in}$ matrix as in layer-wise MoS.
Compared to layer-wise MoS, neuron-wise MoS offers increased flexibility ($m \times {n_{out_a}}$ instead of only $m$) to manage the degree of weight sharing between different architectures, with the number of parameters remaining proportional to $m$. Neuron-wise MoS provides finer control over weight sharing between subnetworks. Gradient conflict, computed using cosine similarity between the supernet and smallest subnet gradients following NASVIT~\citep{gong2021nasvit}, is lowest for neuron-wise MoS compared to layer-wise MoS and HAT, as shown by the highest cosine similarity (see~\ref{sec:grad_confl_sharing}).

\subsection{Adding $g(x,a;E)$ to Transformer}
\label{sec:add_to_trans}
% {MoS is applicable to a single linear layer, multiple linear layers, and other parameterized layers (e.g., layer-norm). Since the linear layer dominates the number of parameters, we follow the approach used in most MoE work~\citep{switch}.} We take the standard weight-sharing based Transformer ($f_a(x;W)$) and replace the two linear layers in every feed-forward network block with $g(x,a;E)$.  % {Experimentally studying the effect of applying MoS on other layers will be our future work.}
MoS is adaptable to a single linear layer, multiple linear layers, and other parameterized layers (e.g., layer-norm). Given that the linear layer dominates the number of parameters, we adopt the approach used in most MoE work~\citep{switch}. We apply MoS to the standard weight-sharing-based Transformer ($f_a(x;W)$) by replacing the two linear layers in every feed-forward network block with $g(x,a;E)$.

%% file: tex/experiments_bert.tex
\section{Experiments - Efficient BERT}
\label{sec:experiments_bert}
%\gan{Experimental setup and results for building efficient BERT models}

\begin{table*}
\scriptsize
\begin{center}
\begin{tabular}% {p{1.2in}p{0.3in}p{0.3in}p{0.3in}p{0.3in}p{0.3in}p{0.3in}p{0.3in}p{0.6in}} 
{cccccccccc}
\toprule
\textbf{Supernet} & \finalchange{\textbf{Training time (hours)}} & \textbf{MNLI} & \textbf{CoLA} & \textbf{MRPC} & \textbf{SST2} & \textbf{QNLI}  & \textbf{QQP}  & \textbf{RTE} & \textbf{Avg. GLUE ($\uparrow$)}     \\ \midrule
Standalone & \finalchange{-} & 82.61 & \textbf{59.03} & 86.54 & 91.52  & 89.47 & 90.68 & 71.53 & 81.63 \\
Supernet (Sandwich) & \finalchange{37} & 82.34 & 57.58 & 86.54 & 91.74 & 88.67 & 90.39 & 73.26 & 81.50 (-0.13) \\
% Arch. Experts Constant Enc(a) (ours)  & 82.07 & 55.32 & \textbf{89.66} & 92.05 & 88.85 & 90.37 & 73.26 & 81.65 (+0.02) \\
Layer-wise MoS (ours) & \finalchange{43 (16\%)} & 82.40 & 57.62 & 87.26 & 92.08 & \textbf{89.57} & \textbf{90.68} & \textbf{77.08} & 82.38 (+0.75) \\
Neuron-wise MoS (ours) & \finalchange{45 (21.6\%)} & \textbf{82.68} & 58.71 & \textbf{87.74} & \textbf{92.16} & 89.22 & 90.49 & 76.39 & \textbf{82.48 (+0.85)}  \\
\bottomrule
\end{tabular}
\caption{GLUE validation performance of different supernets (0 additional pretraining steps) compared to standalone ($1$x pretraining budget). The BERT architecture ($67M$ parameters) is the top model from the pareto front of Supernet (Sandwich) on SuperShaper's search space. Improvement (\%) in GLUE average over standalone is enclosed in parentheses in the last column. Layer-wise and neuron-wise MoS perform significantly better than standalone. \finalchange{For these improvements, MoS imposes a minimal computational overhead of under 22\% for BERT.}}
\label{tab:bert_supernetvsscratch}
\end{center}
\end{table*}

\begin{table*}[t]
\scriptsize
\begin{center}
\begin{tabular}% {p{0.95in}p{0.4in}p{0.55in}p{0.25in}p{0.25in}p{0.25in}p{0.25in}p{0.25in}p{0.25in}p{0.25in}c} 
{ccccccccccc}
\toprule
\textbf{Supernet} & \textbf{\#Params}  & \textbf{\#Steps}  & \textbf{CoLA} & \textbf{MRPC} & \textbf{SST2} & \textbf{QNLI}  & \textbf{QQP}  & \textbf{RTE} & \textbf{Avg. GLUE}     \\ \midrule %  & \textbf{MNLI}
NAS-BERT & 5M & 125K & 19.8 & 79.6 & \textbf{87.3} & 84.9 & 85.8 & 66.7 & {70.7} \\ % 74.4 
AutoDistil (proxy) & 6.88M & 0 &  24.8 & 78.5 & 85.9 & \textbf{86.4} & \textbf{89.1} & 64.3 & {71.5} \\ % \textbf{79.0}
Neuron-wise MoS  & 5M & 0  & \textbf{28.3} & \textbf{82.7} & 86.9 & 84.1 & 88.5 & \textbf{68.1} & {\textbf{73.1}} \\ \midrule % & 75.5
NAS-BERT & 10M & 125K &  34.0 & 79.1 & \textbf{88.6} & \textbf{86.3} & 88.5 & \textbf{66.7} & {73.9} \\ % 76.4 &
Neuron-wise MoS & 10M & 0 &  \textbf{34.7} & \textbf{81.0} & 88.1 & 85.1 & \textbf{89.1} & \textbf{66.7} & {\textbf{74.1}} \\ \midrule % \textbf{77.2} &
AutoDistil (proxy) & 26.1M & 0  & 48.3 & \textbf{88.3} & 90.1 & \textbf{90.0} & 90.6 & 69.4 & {79.5} \\ % & \textbf{83.2}
AutoDistil (agnostic) & 26.8M & 0 &  47.1 & 87.3 & \textbf{90.6} & {89.9} & \textbf{90.8} & 69.0 & {79.1} \\ % {82.8} &
Neuron-wise MoS & 26.8M & 0 & \textbf{52.7} & {88.0} & 90.0 & 87.7 & 89.9 & \textbf{78.1} & {\textbf{81.1}} \\ \midrule %  80.7 &
NAS-BERT & 30M & 125K & 48.7 & 84.6 & 90.5 & \textbf{88.4} & \textbf{90.2} & 71.8 & {{79.0}} \\ % 81.0 & 
Neuron-wise MoS & 30M & 0 &  \textbf{51.0} & \textbf{87.3} & \textbf{91.1} & 87.9 & \textbf{90.2} & \textbf{72.2} & {\textbf{80.0}} \\ \midrule % {\textbf{81.6}} &
AutoDistil (proxy) & 50.1M & 0 & \textbf{55.0} & \textbf{88.8} & 91.1 & \textbf{90.8} & \textbf{91.1} & 71.9 & {81.4} \\ % & \textbf{83.8} 
Neuron-wise MoS & 50M & 0 & \textbf{55.0} & 88.0 & \textbf{91.9} & 89.0 & 90.6 & \textbf{75.4} & {\textbf{81.6}} \\  % & 82.4 
\bottomrule
\end{tabular}
\caption{Comparison of neuron-wise MoS with NAS-BERT and AutoDistil for different model sizes ($\leq50M$ parameters) based on GLUE validation performance. Neuron-wise MoS use a search space of $550$ architectures, which is on par with AutoDistil. The third column corresponds to the number of additional training steps required to obtain the weights for the final architecture after supernet training. Performance numbers for the baseline models are taken from the corresponding papers. {See~\ref{sec:bert_arch_comparison} for the hyperparameters of the best architectures.} On average GLUE, neuron-wise MoS can perform similarly or improves over NAS-BERT for different model sizes without any additional training. Neuron-wise MoS can improve over AutoDistil for most model sizes in average GLUE. }
\label{tab:bert_comparison_to_sota}
\end{center}
\end{table*}

\subsection{Experiment Setup}
We explore the application of our proposed supernet in constructing efficient task-agnostic BERT~\citep{devlin-etal-2019-bert} models, focusing on the BERT pretraining task. This involves pretraining a language model from scratch to learn task-agnostic text representations using a masked language modeling objective. The pretrained BERT model is then finetuned on various downstream NLP tasks. Emphasis is on building highly accurate yet small BERT models (e.g., $5M-50M$ parameters). Both BERT supernet and standalone models are pretrained from scratch on Wikipedia and Books Corpus~\citep{bookscorpus}. Performance evaluation is conducted by finetuning on seven tasks from the GLUE benchmark~\citep{wang-etal-2018-glue}, chosen by AutoDistil~\citep{autodistill}. The {architecture encoding}, data preprocessing, pretraining settings, and finetuning settings are detailed in~\ref{sec:bert_pt_ft_settings}. Baseline models include standalone and standard supernet models proposed in SuperShaper~\citep{supershaper}. Our proposed models are layer-wise and neuron-wise MoS. All supernets undergo sandwich training~\footnote{SuperShaper notes that SPOS performs poorly compared to sandwich training; hence, we do not study SPOS for building BERT models. {The learning curve is shown in~\ref{sec:bert_learning_curve}.}}. Parameters $m$ and router's hidden dimension are set to $2$ and $128$, respectively, for MoS supernets.

\subsection{Supernet vs. standalone gap}
\label{sec:bert_supernet_vs_standalone_gap}

For investigating the supernet vs. standalone gap, the search space is derived from SuperShaper~\citep{supershaper}, encompassing BERT architectures differing only in hidden size at each layer (\textit{{120, 240, 360, 480, 540, 600, 768}}) with fixed numbers of layers ($12$) and attention heads ($12$). This search space includes around $14$ billion architectures. We examine the supernet vs. standalone model gap for the top model architecture from the pareto front of Supernet (Sandwich)~\citep{supershaper}. Table~\ref{tab:bert_supernetvsscratch} illustrates the GLUE benchmark performance of standalone training for the architecture ($1$x pretraining budget, equivalent to 2048 batch size * 125,000 steps) as well as architecture-specific weights from different supernets ($0$ additional pretraining steps; i.e., only supernet pretraining). MoS (layer-wise or neuron-wise) bridges the gap between task-specific supernet and standalone performance for $6$ out of $7$ tasks, including MNLI, a widely used task for evaluating pretrained language models~\citep{liu2019roberta,magic}. The average GLUE gap between the standalone model and standard supernet is $0.13$ points. Remarkably, with customization and expressivity properties, layer-wise and neuron-wise MoS significantly improve standalone training by $0.75$ and $0.85$ average GLUE points, respectively.~\footnote{Consistency of these results across different seeds is discussed in~\ref{sec:bert_diff_seeds}.} \finalchange{Table~\ref{tab:bert_supernetvsscratch} demonstrates that MoS imposes a computational overhead of under 22\% for BERT, resulting in a minimum of 0.8 average GLUE improvement compared to the standard supernet. This overhead may not be significant, as it represents a one-time investment that eliminates the need for additional training after the search process.}

\subsection{Comparison with SoTA NAS}
\label{sec:bert_comparison_with_sota_nas}

The SoTA NAS frameworks for constructing a task-agnostic BERT model are NAS-BERT~\citep{nasbert} and AutoDistil~\citep{autodistill}.\footnote{AutoDistil (proxy) outperforms SoTA distillation approaches such as TinyBERT~\citep{tinybert} and MINILM~\citep{minilm} by $0.7$ average GLUE points. Hence, we do not compare against these works.} % by $0.7$ average GLUE points
The NAS-BERT pipeline comprises: (1) supernet training (with a Transformer stack containing multi-head attention, feed-forward network [FFN], and convolutional layers at arbitrary positions), (2) search based on the distillation (task-agnostic) loss, and (3) pretraining the best architecture from scratch ($1$x pretraining budget, equivalent to 2048 batch size * 125,000 steps). The third step needs to be repeated for every constraint change and hardware change, incurring substantial costs. The AutoDistil pipeline involves: (1) constructing $K$ search spaces and training supernets for each search space independently, (2a) \textit{agnostic}-search mode: searching based on the self-attention distillation (task-agnostic) loss, (2b) \textit{proxy}-search mode: searching based on the MNLI validation score, and (3) extracting architecture-specific weights from the supernet without additional training. The first step can be costly as pretraining $K$ supernets requires $K$ times training compute and memory, compared to training a single supernet. The proxy-search mode may favor AutoDistil unfairly, as it finetunes all architectures in its search space on MNLI and utilizes the MNLI score for ranking. {For a fair comparison with SoTA, MNLI task is excluded from evaluation.}%~\footnote{See~\ref{sec:bert_arch_comparison_incl_mnli} for the comparison of neuron-wise MoS against baselines that do not directly tune on the MNLI task, where we find that neuron-wise MoS improves over baselines consistently in terms of both average GLUE and MNLI task performance.}
~\footnote{Refer to~\ref{sec:bert_arch_comparison_incl_mnli} for a comparison of neuron-wise MoS against baselines that don't directly tune on the MNLI task. Neuron-wise MoS consistently outperforms baselines in terms of both average GLUE and MNLI task performance.}

Our proposed NAS pipeline addresses challenges in NAS-BERT and AutoDistil. In comparison to SoTA NAS, our search space includes BERT architectures with uniform Transformer layers: hidden size ($120$ to $768$ in increments of $12$), attention heads (\textit{{6, 12}}), intermediate FFN hidden dimension ratio (\textit{{2, 2.5, 3, 3.5, 4}}). This search space comprises $550$ architectures, similar to AutoDistil. The supernet is based on neuron-wise MoS, and the search uses perplexity (task-agnostic) to rank architectures. Unlike NAS-BERT, our final architecture weights are directly extracted from the supernet without additional pretraining. Unlike AutoDistil, our pipeline pretrains only one supernet, significantly reducing training compute and memory. We use only task-agnostic metrics for search, similar to AutoDistil's agnostic setting.
Table~\ref{tab:bert_comparison_to_sota} compares neuron-wise MoS supernet with NAS-BERT and AutoDistil for various model sizes. NAS-BERT and AutoDistil performances are obtained from respective papers. On average GLUE, our pipeline improves over NAS-BERT for $5M$, $10M$, and $30M$ model sizes, with no additional training (100\% additional training compute savings, equivalent to 2048 batch size * 125,000 steps). On average GLUE, our pipeline: (i) surpasses AutoDistil-proxy for $6.88M$ { and $50M$ model sizes} with $1.88M$ and $0.1M$ fewer parameters respectively, and (ii) outperforms both AutoDistil-proxy and AutoDistil-agnostic for $26M$ model size. Besides achieving SoTA results, our method significantly reduces the heavy workload of training multiple models in subnetwork retraining (NAS-BERT) or supernet training (AutoDistil).

%% file: tex/experiments_mt.tex
\section{Experiments - Efficient MT}
\label{sec:experiments_mt}
%\gan{Experimental setup and results for building efficient machine translation models}
% In this section, we discuss the application of proposed supernets for building efficient MT models. 

\begin{table*}
\scriptsize
\begin{center}
\begin{tabular}%{p{1.0in}p{0.55in}p{0.55in}p{0.55in}p{0.55in}p{0.55in}p{0.55in}} 
{ccccccc}
\toprule
\textbf{Dataset} & \multicolumn{2}{c}{\textbf{WMT'14 En-De}} & \multicolumn{2}{c}{\textbf{WMT'14 En-Fr}} & \multicolumn{2}{c}{\textbf{WMT'19 En-De}} \\
\textbf{Supernet} & \textbf{MAE ($\downarrow$)} & \textbf{Kendall ($\uparrow$)} & \textbf{MAE ($\downarrow$)} & \textbf{Kendall ($\uparrow$)} & \textbf{MAE ($\downarrow$)} & \textbf{Kendall ($\uparrow$)} 
\\ \midrule
HAT  & 1.84 & 0.81 & 1.37 & 0.63 & 2.07 & 0.71 \\
Supernet (Sandwich) & 1.62 (12\%) & 0.81 & 1.37 (0\%) & 0.63 & 2.02 (2.4\%) & 0.87 \\
Layer-wise MoS (ours) & 1.61 (12.5\%) & 0.54 & 1.24 (9.5\%) & 0.73 & 1.57 (24.2\%) & 0.87 \\
Neuron-wise MoS (ours) & \textbf{1.13 (38.6\%)} & 0.71 & \textbf{1.2 (12.4\%)} & 0.85 & \textbf{1.48 (28.5\%)} & 0.81 \\
\bottomrule
\end{tabular}
\caption{Mean absolute error (MAE) and Kendall rank correlation coefficient between the supernet and the standalone model BLEU performance for 15 random architectures from the MT search space. Improvements (\%) in mean absolute error over HAT are in parentheses. Our supernets enjoy minimal MAE and comparable ranking quality with respect to the baseline models.}
\label{tab:mt_supernetvsscratch}
\end{center}
\end{table*}

\subsection{Experiment setup}

We discuss the application of proposed supernets for building efficient MT models following the setup by Hardware-aware Transformers (HAT~\citep{wang-etal-2020-hat}), the SoTA NAS framework for MT models with good latency-BLEU tradeoffs. Focusing on WMT'14 En-De, WMT'14 En-Fr, and WMT'19 En-De benchmarks, we maintain consistent architecture encoding and training settings for supernet and standalone models (details in~\ref{sec:mt_train_settings}). Baseline supernets include HAT and Supernet (Sandwich). Proposed supernets are Layer-wise MoS and Neuron-wise MoS, both using sandwich training, with $m$ and router's hidden dimension set to 2 and 128, respectively. {Refer to~\ref{sec:mt_vary_m} for the rationale behind choosing `$m$'.}

\subsection{Supernet vs. standalone gap}
\label{sec:mt_super_stand_gap}

In HAT's search space of $6M$ encoder-decoder architectures, featuring flexible parameters like embedding size (512 or 640), decoder layers (1 to 6), self/cross attention heads (4 or 8), and number of top encoder layers for decoder attention (1 to 3), good supernets should exhibit minimal mean absolute error (MAE) and high rank correlation between supernet and standalone performance for a given architecture. Table~\ref{tab:mt_supernetvsscratch} presents MAE and Kendall rank correlation for 15 random architectures, showcasing that sandwich training yields better MAE and rank quality compared to HAT. While our proposed supernets achieve comparable rank quality for WMT'14 En-Fr and WMT'19 En-De, and slightly underperform for WMT'14 En-De, they exhibit minimal MAE across all tasks. Particularly, neuron-wise MoS achieves substantial MAE improvements, suggesting lower additional training steps needed to make MAE negligible (as detailed in Section~\ref{sec:mt_addtrain}). Supernet and standalone performance plots reveal neuron-wise MoS excelling for almost all top-performing architectures (see~\ref{sec:mt_supernet_vs_standalone}). The training overhead for MoS is generally negligible, e.g., for WMT'14 En-De, supernet training takes 248 hours, with neuron-wise MoS and layer-wise MoS requiring 14 and 18 additional hours, respectively (less than $8$\% overhead, {see Section~\ref{sec:mt_overall_time_breakdown} for details}).

\begin{comment}
\begin{figure*}[t!]
    \centering
    \begin{subfigure}[t]{0.3\textwidth}
        \centering
        \includegraphics[height=1.3in, width=2.15in]{images/mt/comparewsota/wmt14.en-de.png}
        \caption{WMT'14 En-De}
    \end{subfigure}
    ~ 
    \begin{subfigure}[t]{0.3\textwidth}
        \centering
        \includegraphics[height=1.3in, width=2.15in]{images/mt/comparewsota/wmt14.en-fr.png}
        \caption{WMT'14 En-Fr}
    \end{subfigure}
    ~ 
    \begin{subfigure}[t]{0.3\textwidth}
        \centering
        \includegraphics[height=1.3in, width=2.15in]{images/mt/comparewsota/wmt19.en-de.png}
        \caption{WMT'19 En-De}
    \end{subfigure}
    \caption{Latency and BLEU trade-offs on the NVIDIA V100 GPU.}
    \label{fig:mt_comparewithsota}
\end{figure*}
\end{comment}

\begin{table*}[t]
\scriptsize
\begin{center}
\begin{tabular}%{p{1.3in}p{0.3in}p{0.3in}p{0.3in}p{0.3in}p{0.3in}p{0.3in}p{0.3in}p{0.3in}p{0.3in}} 
{cccccccccc}
\toprule
\textbf{Dataset} & \multicolumn{3}{c}{\textbf{WMT'14 En-De}} & \multicolumn{3}{c}{\textbf{WMT'14 En-Fr}} & \multicolumn{3}{c}{\textbf{WMT'19 En-De}} \\
\textbf{Supernet / Latency Constraint} & \textbf{100 ms} & \textbf{150 ms} & \textbf{200 ms} & \textbf{100 ms} & \textbf{150 ms} & \textbf{200 ms} & \textbf{100 ms} & \textbf{150 ms} & \textbf{200 ms}
\\ \midrule
HAT & 25.26 & 26.25 & 26.28 & 38.94 & 39.26 & 39.16 & 42.61 & 43.07 & 43.23 \\
Layer-wise MoS (ours) & 26.28 & 27.31 & \textbf{28.03} & 39.34 & \textbf{40.29} & \textbf{41.24} & 43.45 & \textbf{44.71} & 46.18 \\
Neuron-wise MoS (ours) & \textbf{26.37} & \textbf{27.59} & 27.79 & \textbf{39.55} & 40.02 & 41.04 & \textbf{43.77} & 44.66 & \textbf{46.21} \\
\bottomrule
\end{tabular}
\caption{Latency vs. Supernet BLEU for the models on the pareto front, obtained by performing search with different latency constraints ($100$ ms, $150$ ms, $200$ ms) on the NVIDIA V100 GPU. Our supernets yield architectures that enjoy better latency-BLEU tradeoffs than HAT.}
\label{tab:mt_comparewithsota}
\end{center}
\end{table*}

\subsection{Comparison with the SoTA NAS}

The pareto front from the supernet is obtained using an evolutionary search algorithm that leverages the supernet for quickly identifying top-performing candidate architectures and a latency estimator for promptly discarding candidates with latencies surpassing a threshold. Settings for the evolutionary search algorithm and latency estimator are detailed in~\ref{sec:hat_settings}. Three latency thresholds are explored: $100$ ms, $150$ ms, and $200$ ms. Table~\ref{tab:mt_comparewithsota} illustrates the latency vs. supernet performance tradeoff for models in the pareto front from different supernets. Compared to HAT, the proposed supernets consistently achieve significantly higher BLEU for each latency threshold across all datasets, emphasizing the importance of architecture specialization and expressiveness of the supernet. {See~\ref{sec:evosearch_mt_stability} for the consistency of these trends across different seeds.}

\begin{table*}
\scriptsize
\begin{center}
\begin{tabular}{p{0.43in}p{0.65in}p{0.65in}p{0.65in}p{0.65in}p{0.65in}p{0.65in}} 
% {ccccccc}
\toprule
\textbf{Dataset} & \multicolumn{3}{c}{\textbf{Additional training steps ($\downarrow$)}} & \multicolumn{3}{c}{{\textbf{Additional training time (NVIDIA V100 hours) ($\downarrow$)}}}  \\
\textbf{Supernet} & \textbf{WMT'14 En-De} & \textbf{WMT'14 En-Fr} & \textbf{WMT'19 En-De} & \textbf{WMT'14 En-De} & \textbf{WMT'14 En-Fr} & \textbf{WMT'19 En-De} \\ \midrule
HAT & 33K & 33K & 26K & 63.9 & \textbf{60.1} & 52.3 \\
Laye. MoS & 16K (51.5\%) & 30K (9\%) & 20K (23\%) & 35.5 (44.4\%)  & 66.5 (-10.6\%) & 45.2 (13.5\%) \\
Neur. MoS & \textbf{13K (60\%)}  & \textbf{26K (21\%)} & \textbf{16K (38.4\%)} & \textbf{31.0 (51.4\%)} & 61.7 (-2.7\%) & \textbf{39.5 (24.5\%)} \\
\bottomrule
\end{tabular}
\caption{Average number of additional training steps and time required for the models on the pareto front to close the supernet vs. standalone gap. Improvements (\%) over HAT are shown in parentheses. Our supernets require minimal number of additional training steps and time to close the gap compared to HAT for most tasks. {See~\ref{sec:addtrain_2data} for each latency constraint.}
\vspace{-1em}
}
\label{tab:mt_addtrainsteps}
\end{center}
\end{table*}

\subsection{Additional training to close the gap}
\label{sec:mt_addtrain}

The proposed supernets significantly minimize the gap between the supernet and standalone MAE (as discussed in Section~\ref{sec:mt_super_stand_gap}), yet the gap remains non-negligible. Closing the gap for an architecture involves extracting architecture-specific weights from the supernet and conducting additional training until the standalone performance is reached (achieving a gap of $0$). An effective supernet should demand a minimal number of additional steps {and time} for the extracted architectures to close the gap. In the context of additional training, we evaluate the test BLEU for each architecture after every $10K$ steps, stopping when the test BLEU matches or exceeds the test BLEU of the standalone model. Table~\ref{tab:mt_addtrainsteps} presents the average number of additional training steps required for all models on the pareto front from each supernet to close the gap. {Compared to HAT, layer-wise MoS achieves an impressive reduction of $9$\% to $51$\% in training steps, while neuron-wise MoS delivers the most substantial reduction of $21$\% to $60$\%. For the WMT'14 En-Fr task, both MoS supernets require at least $2.7$\% more time than HAT to achieve SoTA BLEU across different constraints. These results underscore the importance of architecture specialization and supernet expressivity in significantly improving the training efficiency of subnets extracted from the supernet.}

\subsection{Comparison to AutoMoE}
\label{sec:automoe_comparison}
\begin{table*}[t]
\scriptsize
\begin{center}
\begin{tabular}%{p{1.3in}p{0.3in}p{0.3in}p{0.3in}p{0.3in}p{0.3in}p{0.3in}p{0.3in}p{0.3in}p{0.3in}} 
{cccccccccc}
\toprule
\textbf{Dataset} & \multicolumn{1}{c}{\textbf{WMT'14 En-De}} & \multicolumn{1}{c}{\textbf{WMT'14 En-Fr}} & \multicolumn{1}{c}{\textbf{WMT'19 En-De}} \\
\textbf{Supernet / Latency Constraint} & \textbf{200 ms} &  \textbf{200 ms}  & \textbf{200 ms}
\\ \midrule
HAT & 26.28 & 39.16 & 43.23 \\
AutoMoE & 26.06 & 38.98 & 43.13 \\
Layer-wise MoS (ours) & \textbf{28.03} & \textbf{41.24} & 46.18 \\
Neuron-wise MoS (ours) & 27.79 & 41.04 & \textbf{46.21} \\
\bottomrule
\end{tabular}
\caption{Latency vs. Supernet BLEU for the models on the pareto front, obtained by performing search with latency constraint of $200$ ms on the NVIDIA V100 GPU. Our supernets yield architectures that enjoy better latency-BLEU tradeoffs than AutoMoE.}
\label{tab:automoe_comparison}
\end{center}
\end{table*}

\finalchange{Although AutoMoE~\cite{jawahar-etal-2023-automoe} and MoS pursue distinct objectives (as discussed in Appendix~\ref{sec:c4_comparison_to_automoe}), we proceed to compare the supernet BLEU scores of HAT, AutoMoE, and MoS under a latency constraint of 200 ms on the NVIDIA V100 GPU across the three WMT benchmarks. Table~\ref{tab:automoe_comparison} shows that MoS consistently outperforms AutoMoE and HAT across all datasets. Interestingly, AutoMoE falls behind HAT, suggesting a potential discrepancy between the performance of AutoMoE's supernet and standalone models.}

% impact of fixed arch. encoding. #122

% plotting alpha's

%% file: tex/related.tex
\section{Related Work}
\label{sec:related_work}

In this section, we briefly review existing research on NAS in NLP. Evolved Transformer (ET)~\citep{evotrans} is an initial work that explores NAS for efficient MT models. It employs evolutionary search and dynamically allocating training resources for promising candidates., HAT~\citep{wang-etal-2020-hat} introduces a weight-sharing supernet as a performance estimator, amortizing the training cost for candidate MT evaluations in evolutionary search. NAS-BERT~\citep{nasbert} partitions the BERT-Base model into blocks and trains a weight-sharing supernet to distill each block. NAS-BERT uses progressive shrinking during supernet training to prune less promising candidates, identifying top architectures for each efficiency constraint quickly. However, NAS-BERT requires pretraining the top architecture from scratch for every constraint change, incurring high costs. SuperShaper~\citep{supershaper} pretrains a weight-sharing supernet for BERT using a masked language modeling objective with sandwich training. AutoDistil~\citep{autodistill} employs few-shot NAS~\citep{fewshotnas}: constructing $K$ search spaces of non-overlapping BERT architectures and training a weight-sharing BERT supernet for each search space. The search is based on self-attention distillation loss with BERT-Base (task-agnostic search) and MNLI score (proxy search). \update{AutoMoE~\citep{jawahar-etal-2023-automoe} augments the search space of HAT with mixture-of-expert models to design efficient translation models. Refer to~\ref{sec:c4_comparison_to_automoe} for the main differences between our framework and the AutoMoE framework.}

%In computer vision community, K-shot NAS~\citep{kshotnas} generates the weight for each subnet as a convex combination of different supernet weights in a dictionary with a simplex code. Their framework is similar to layer-wise MoS with the following key differences. K-shot NAS trains the architecture code generator and supernet iteratively due to training difficulty, while layer-wise MoS trains all its components jointly. K-shot NAS has been applied only in convolutional architectures for image classification tasks. K-shot NAS introduces too many parameters with increase in number of supernets ($K$), which is alleviated by neuron-wise MoS due to its granular weight specialization. In this work, we focus on tasks in NLP (and the relevant baselines), where we find that the supernets lag behind standalone models significantly in terms of performance. Also, authors of k-shot NAS do not release the code to reproduce their results. Hence, we do not evaluate against k-shot NAS.
% Primer~\cite{primer} searches for training-efficient decoder-only autoregressive models from a search space containing TensorFlow primitives. They use evolutionary search to identify the pareto front, where each candidate architecture is trained from scratch for 24 TPUv2 hours. The best architecture matches the one-shot performance of GPT-3 XL~\cite{gpt3} at 1/3 of the training compute. LiteTransformerSearch~\cite{litetransformersearch} uses the number of parameters in the decoder as a training-free performance proxy metric to search for efficient autoregressive models. 

In the computer vision community, K-shot NAS~\citep{kshotnas} generates weights for each subnet as a convex combination of different supernet weights in a dictionary using a simplex code. While K-shot NAS shares similarities with layer-wise MoS, there are key distinctions. K-shot NAS trains the architecture code generator and supernet iteratively due to training difficulty, whereas layer-wise MoS trains all its components jointly. K-shot NAS has been applied specifically to convolutional architectures for image classification tasks. However, it introduces too many parameters with an increase in the number of supernets ($K$), a concern alleviated by neuron-wise MoS due to its granular weight specialization. In our work, we focus on NLP tasks and relevant baselines, noting that supernets in NLP tend to lag significantly behind standalone models in terms of performance. Additionally, the authors of K-shot NAS have not released the code to reproduce their results, preventing a direct evaluation against their method.

%\subsection{Efficient BERT}
%\gan{Summarize: NAS-BERT, SuperShaper, MAGIC, AutoDistil, AutotinyBERT, ...}

%\subsection{Efficient Machine Translation}
%\gan{Summarize: HAT, Evolved Transformers,}

%% file: tex/conclusion.tex
\section{Conclusion}
\label{sec:conclusion}

We introduced Mixture-of-Supernets, a formulation aimed at enhancing the expressive power of supernets. By adopting the idea of MoE, we demonstrated the ability to generate flexible weights for subnetworks. Through extensive evaluations for constructing efficient BERT and MT models, our supernets showcased the capacity to: (i) minimize retraining time, thereby significantly improving NAS efficiency, and (ii) produce high-quality architectures that meet user-defined constraints.

%% file: tex/limitations.tex
\section{Limitations}
\label{sec:limitations}
%\gan{limitations: i) no scratch vs supernet plot for random archs - BERT (expensive), ii) M=2 only }
The limitations of this work are as follows:
\begin{enumerate}% [leftmargin=*,noitemsep,topsep=0pt,parsep=0pt,partopsep=0pt]
    \item Applying Mixture-of-Supernet (MoS) to popular benchmarks in NLP, focusing on efficient machine translation and BERT, offers valuable insights. A potential impactful future direction could involve extending the application of MoS to build efficient autoregressive decoder-only language models, such as GPT-4~\cite{openai2023gpt4}. % We study the application of Mixture-of-Supernet (MoS) only on the popular NAS benchmarks for NLP, which include efficient machine translation and BERT benchmarks. Given the growing interest in building autoregressive decoder-only language models such as GPT-4~\cite{openai2023gpt4}, applying MoS to build efficient GPT model will be an impactful future direction.     
    % \item We apply the proposed generalized model function only on the linear layers in the feed-forward network block. It would be interesting to study the application of the function on other Transformer components such as self-attention projection layers and LayerNorm. \hcnotes{we need have a reason why we didn't try it now. Since this does not sound a lot of work, maybe not a good limitation to write.}    \hcnotes{one thing we can say here is we didn't apply our method on more NLP tasks. In this paper we focus on the tasks where some other NAS works also tried for comparison. We will try the proposed method on other NLP tasks such as XXX, etc. in the future work.}
    \item  Introducing MoE architecture potentially need more training budget. In our work, we do not use large number of training iteration for fair comparison and fixing the number of expert weights ($m$) to $2$ works well. We will investigate the full potential of the proposed supernets by combining larger training budget (e.g., $\geq 200K$ steps) and larger number of expert weights (e.g., $\geq 16$ expert weights) in the future work. %in this work. In our initial experiments, we find that the supernet performance to drop steadily as $m$ is increased from $3$ for the standard training budget ($40K$ steps for MT and $125K$ steps for BERT). It would be interesting to scale both $m$ and the number of training steps jointly to study the impact of larger supernets on the search quality. 
    % \hcnotes{we need have a reason why we didn't try it now. We can say that introudcing MoE architecture potentially need more training budget. In the current work we does not use larger number of training iteration for fair comparison. we will investigate the full potential of the proposed method by combining larger training budget in the future work.}
    \item % Due to high computational requirements for pretraining BERT, we study the supernet vs. the standalone model gap only for the top model from the pareto front of Supernet (Sandwich) (see Table~\ref{tab:bert_supernetvsscratch}). It would be interesting to study the gap for a large number of architectures from the search space (similar to Table~\ref{tab:mt_supernetvsscratch}).
    \finalchange{Due to the high computational requirements for pretraining BERT, we only investigate the gap between the supernet and standalone models for the top model from the pareto front of the Supernet (Sandwich) (see Table~\ref{tab:bert_supernetvsscratch}). It would be interesting to explore this gap for a larger number of architectures from the search space, as shown in Table~\ref{tab:mt_supernetvsscratch} for MT tasks.}
\end{enumerate}

\section*{Acknowledgments}
\label{sec:ack}
\finalchange{MAM acknowledges support from Canada Research Chairs (CRC), the Natural Sciences and Engineering Research Council of Canada (NSERC; RGPIN-2018-04267), Canadian Foundation for Innovation (CFI; 37771), and Digital Research Alliance of Canada.\footnote{\href{https://alliancecan.ca}{https://alliancecan.ca}} Lakshmanan's research was supported in part by a grant from NSERC (Canada).} We used ChatGPT for rephrasing and grammar checking of the paper.

%% file: tex/appendix.tex
\section{Appendix}
\label{sec:appendix}

\subsection{Weight Sharing and Gradient Conflict Analysis}

\subsubsection{Jensen-Shannon distance of alignment vector as a weight sharing measure}
\label{sec:layer_wise_sharing}
\begin{table*}
\scriptsize
\begin{center}
\begin{tabular}{ccccc} \toprule
\textbf{Model 1} & \textbf{Model 2}  & \textbf{WMT'14 En-De}  & \textbf{WMT'14 En-Fr} & \textbf{WMT'19 En-De} \\
\midrule
Smallest A ($23M$) & Largest A ($118M$) & 0.297 & 0.275 & 0.263 \\
Smallest B ($23M$) & Largest B ($118M$) & 0.281 & 0.258 & 0.245 \\
Smallest A ($23M$) & Largest B ($118M$) & 0.284 & 0.263 & 0.249 \\
Smallest B ($23M$) & Largest A ($118M$) & 0.294 & 0.27 & 0.259 \\
Smallest A ($23M$) & Smallest B ($118M$) & 0.006 & 0.008 & 0.004 \\
Largest A ($23M$) & Largest B ($118M$) & 0.014 & 0.012 & 0.015 \\
\bottomrule
\end{tabular}
\caption{{Jensen-Shannon distance of Layer-wise MoS alignment vector across models as a weight sharing measure. Layer-wise MoS induces low degree of weight sharing between differently sized architectures shown by higher Jensen-Shannon distance between their alignment vectors compared to that of similarly sized architectures. Note that architectures A and B differ by number of encoder/decoder attention heads.}}
\label{tab:layer_wise_sharing}
\end{center}
\end{table*}
{We use the Jensen-Shannon distance of alignment vector generated by Layer-wise MoS for two architectures as a proxy to quantify the degree of weight sharing. Ideally, the lower the Jensen-Shannon distance, the higher the degree of weight sharing and vice-versa. We experiment with two architectures of 23M parameters (Smallest A and Smallest B) and two architectures of 118M parameters (Largest A and Largest B). From Table~\ref{tab:layer_wise_sharing}, it is clear that Layer-wise MoS induces low degree of weight sharing between differently sized architectures shown by higher Jensen-Shannon distance between their alignment vectors. On the other hand, there is a high degree of weight sharing between similarly sized architectures where Jensen-Shannon distance is significantly low.}

\subsubsection{Cosine similarity between the supernet gradient and the smallest subnet gradient as a gradient conflict measure.}
\label{sec:grad_confl_sharing}
\begin{table*}
\scriptsize
\begin{center}
\begin{tabular}{ccc} \toprule
\textbf{Supernet} & \textbf{WMT'14 En-De}  & \textbf{WMT'19 En-De}  \\
\midrule
HAT & 0.522 & 0.416 \\
Layer-wise MoS & 0.515 & 0.517 \\
Neuron-wise MoS & 0.555 & 0.52 \\
\bottomrule
\end{tabular}
\caption{{Gradient conflict via cosine similarity between the supernet gradient and the smallest subnet gradient. Neuron-wise MoS enjoys lower gradient conflict, shown via. high cosine similarity. }}
\label{tab:grad_confl_sharing}
\end{center}
\end{table*}
{We compute gradient conflict using cosine similarity between the supernet gradient and the smallest subnet gradient, following NASVIT work~\citep{gong2021nasvit}. In Table~\ref{tab:grad_confl_sharing}, we show that Neuron-wise MoS enjoys lowest gradient conflict compared to Layer-wise MoS and HAT, shown by highest cosine similarity.}

\subsection{Detailed algorithm for Supernet training and Search}
\label{sec:detailed_algorithm}

\subsubsection{Supernet training algorithm}
\label{sec:algo_supernet_training}
{The detailed algorithm for supernet training is shown in Algorithm~\ref{algo:supernet_training}.}

\begin{algorithm*}
\hspace*{\algorithmicindent} \textbf{Input:} $\texttt{\mbox{Training data:} $\mathcal{X}_{tr}$}$, $\texttt{\mbox{Search space:} $\mathcal{A}$}$, \\ $\texttt{\mbox{No. of training steps:} \mbox{num-train-steps}}$, $\texttt{\mbox{Type of MoS:} \mbox{mos-type}}$ \\
     \hspace*{\algorithmicindent} \textbf{Output:} $\texttt{\mbox{Training Supernet Weights:} $\mathbb{E}$}$
     \begin{algorithmic}[1]
     \State $\mathbb{E} \gets$ Random weights from Normal Distribution.
     \For{$iter \gets 1 \: \texttt{\mbox{to}} \: \texttt{\mbox{num-train-steps}}$}
        \State {\textcolor{blue}{// sample data}}
        \State $x, y \sim \mathcal{X}_{tr}$
        \State {\textcolor{blue}{// perform sandwich sampling}}
        \For{a in [$ a_{rand} \sim \mathcal{A}, a_{big}, a_{small} $]}
            \State \text{Enc}(a) {\textcolor{blue}{// create the architecture encoding}}
           \State {\textcolor{blue}{// generate architecture-specific weights}}
            \If{\mbox{mos-type} == \mbox{Layer wise MoS}}
               \State $W_a = \sum_i r(\text{Enc}(a))^i  E^i_a $
            \ElsIf{\mbox{mos-type} == \mbox{Neuron wise MoS}}
               \State $W_a = \sum_i \text{diag}(\beta^i_a) E^i_a$
            \EndIf
           \State {\textcolor{blue}{// compute task-specific loss}}
           \State $\mbox{loss} \gets \mathcal{L}(W_ax,y)$
           \State $\mbox{loss}.$backward() {\textcolor{blue}{// compute gradients}}
        \EndFor 
        \State Update $\mathbb{E}$ using accumulated gradients {\textcolor{blue}{// learning rule}}
     \EndFor
     \State return $\mathbb{E}$
\end{algorithmic}
\caption{Training algorithm for Mixture-of-Supernets used in MT.}
\label{algo:supernet_training}
\end{algorithm*}

\subsubsection{Search algorithm}
\label{sec:algo_search}
{The detailed algorithm for search is shown in Algorithm~\ref{algo:evosearch}.}

\begin{algorithm*}
\hspace*{\algorithmicindent} \textbf{Input:} $\texttt{\mbox{supernet}}$, $\texttt{\mbox{latency-predictor}}$, $\texttt{\mbox{num-iterations}}$, $\texttt{\mbox{num-population}}$, $\texttt{\mbox{num-parents}}$, $\texttt{\mbox{num-mutations}}$, $\texttt{\mbox{num-crossover}}$, $\texttt{\mbox{mutate-prob}}$, $\texttt{\mbox{latency-constraint}}$ \\
     \hspace*{\algorithmicindent} \textbf{Output:} $\texttt{\mbox{best-architecture}}$ 
     \begin{algorithmic}[1]
     \State {\textcolor{blue}{// create initial population}}
     \State $popu \gets$ $\texttt{\mbox{num-population}}$ random samples from the search space
     \For{$iter \gets 1 \: \texttt{\mbox{to}} \: \texttt{\mbox{num-iterations}}$}
        \State {\textcolor{blue}{// generate parents by picking top candidates}}
        \State {\textcolor{red}{$\texttt{\mbox{cur-parents}} \gets$ top `$\texttt{\mbox{num-parents}}$' architectures from $popu$ by MoS validation loss}}
        \State {\textcolor{blue}{// generate candidates via mutation}}
        \State $\texttt{\mbox{cur-mutate-popu}}$ = $\{ \}$
        \For{$mi \gets 1 \: \texttt{\mbox{to}} \: \texttt{\mbox{num-mutations}}$}
            \State $\texttt{\mbox{cur-mutate-gene}} \gets$ mutate a random example from $popu$ with mutation probability $\texttt{\mbox{mutate-prob}}$
            \If{$\texttt{\mbox{cur-mutate-gene}}$ satisfies $\texttt{\mbox{latency-constraint}}$ via $\texttt{\mbox{latency-predictor}}$}
               \State $\texttt{\mbox{cur-mutate-popu}} = \texttt{\mbox{cur-mutate-popu}} \cup \texttt{\mbox{cur-mutate-gene}}$
            \EndIf
        \EndFor
        \State {\textcolor{blue}{// generate candidates via cross-over}}
        \State $\texttt{\mbox{cur-crossover-popu}}$ = $\{ \}$
        \For{$ci \gets 1 \: \texttt{\mbox{to}} \: \texttt{\mbox{num-crossover}}$}
            \State $\texttt{\mbox{cur-crossover-gene}} \gets$ crossover two random examples from $popu$
            \If{$\texttt{\mbox{cur-crossover-gene}}$ satisfies $\texttt{\mbox{latency-constraint}}$ via $\texttt{\mbox{latency-predictor}}$}
               \State $\texttt{\mbox{cur-crossover-popu}} = \texttt{\mbox{cur-crossover-popu}} \cup \texttt{\mbox{cur-crossover-gene}}$
            \EndIf
        \EndFor
        \State {\textcolor{blue}{// update population}}
        \State $popu = \texttt{\mbox{cur-parents}} \cup \texttt{\mbox{cur-mutate-popu}} \cup \texttt{\mbox{cur-crossover-popu}}$
     \EndFor
     \State {\textcolor{red}{return top architecture from $popu$ by MoS's validation loss}}
\end{algorithmic}
\caption{Evolutionary search algorithm for Neural architecture search used in MT.}
\label{algo:evosearch}
\end{algorithm*}

\subsection{Comparison to the AutoMoE work}
\label{sec:c4_comparison_to_automoe}

\noindent\textbf{{Goals}}:
\update{Given a search space of dense and mixture-of-expert models, the goal of the AutoMoE framework~\cite{jawahar-etal-2023-automoe} is to search for high-performing model architectures that satisfy user-defined efficiency constraints. The final architectures can be dense or mixture-of-expert models. On the other hand, given a search space of dense models only, the goal of the Mixture-of-Supernets framework is to search for high-performing dense model architectures that satisfy user-defined efficiency constraints. The final architecture can be a dense model only. In addition, the MoS framework minimizes the retraining compute required for the searched architecture to approach the standalone performance. The MoS framework designs the supernet with  flexible weight sharing and high capacity. On the other hand, the supernet underlying the AutoMoE framework suffers from strict weight sharing and limited capacity.}

\noindent\textbf{{Applications of mixture-of-experts}}:
\update{The main application of mixture-of-experts idea by the AutoMoE framework is to augment the standard NAS search space of dense models with mixture-of-experts models. To this end, the AutoMoE framework modifies the standard weight sharing supernet to support weight generation for mixture-of-expert models. On the other hand, the Mixture-of-Supernets framework uses the mixture-of-expert design to: (i) increase the capacity of standard weight sharing supernet and (ii) customize weights for each architecture. Post training, the expert weights are collapsed to create a single weight for the dense architecture.} 

\noindent\textbf{{Router specifications}}:
\update{The router underlying the AutoMoE framework takes token embedding as input, outputs a probability distribution over experts, and passes token embedding to top-k experts. On the other hand, the router underlying the Mixture-of-Supernets framework takes architecture embedding as input, outputs a probability distribution over experts (layer-wise MoS) / neurons (neuron-wise MoS), uses the probability distribution to combine ALL the expert weights into a single weight, and passes token embedding to the single weight (all experts).}

\subsection{Additional Experiments - Efficient BERT}
\label{sec:appendix_experiments_bert}

\subsubsection{BERT pretraining / finetuning settings}
\label{sec:bert_pt_ft_settings}
\noindent\textbf{\underline{Pretraining data}}: The pretraining data consists of text from Wikipedia and Books Corpus~\citep{bookscorpus}. We use the data preprocessing scripts provided by \citeauthor{izsak-etal-2021-train} to construct the tokenized text.

\noindent\textbf{\underline{Supernet and standalone pretraining settings}}: The pretraining settings for supernet and standalone models are taken from SuperShaper~\citep{supershaper}: batch size of 2048, maximum sequence length of 128, training steps of 125K, learning rate of $5e^{-4}$, weight decay of $0.01$, and warmup steps of $10K$ ($0$ for standalone). {For experiments with the search space from SuperShaper~\citep{supershaper} (Section~\ref{sec:bert_supernet_vs_standalone_gap}), the architecture encoding $a$ is a list of hidden size at each layer of the architecture (12 elements since the supernet is a 12 layer model). For experiments with the search space on par with AutoDistil~\citep{autodistill} (Section~\ref{sec:bert_comparison_with_sota_nas}), the architecture encoding $a$ is a list of four elastic hyperparameters of the homogeneous BERT architecture: number of layers, hidden size of all layers, feedforward network (FFN) expansion ratio of all layers and number of attention heads of all layers (see Table~\ref{tab:bert_arch_analysis_autodistil} for sample homogeneous BERT architectures).}

\noindent\textbf{\underline{Finetuning settings}}: We evaluate the performance of the BERT model by finetuning on each of the seven tasks (chosen by AutoDistil~\citep{autodistill}) in the GLUE benchmark~\citep{wang-etal-2018-glue}. The evaluation metric is the average accuracy (Matthews's correlation coefficient for CoLA only) on all the tasks (GLUE average). The finetuning settings are taken from the BERT paper~\citep{devlin-etal-2019-bert}: learning rate from \{$5e^{-5}$, $3e^{-5}$, $2e^{-5}$\}, batch size from \{16, 32\}, and epochs from \{2, 3, 4\}.

\subsubsection{Learning curve for BERT supernet variants}
\label{sec:bert_learning_curve}
\begin{figure*}
\centering
\includegraphics[width=4.5in,height=2.5in]{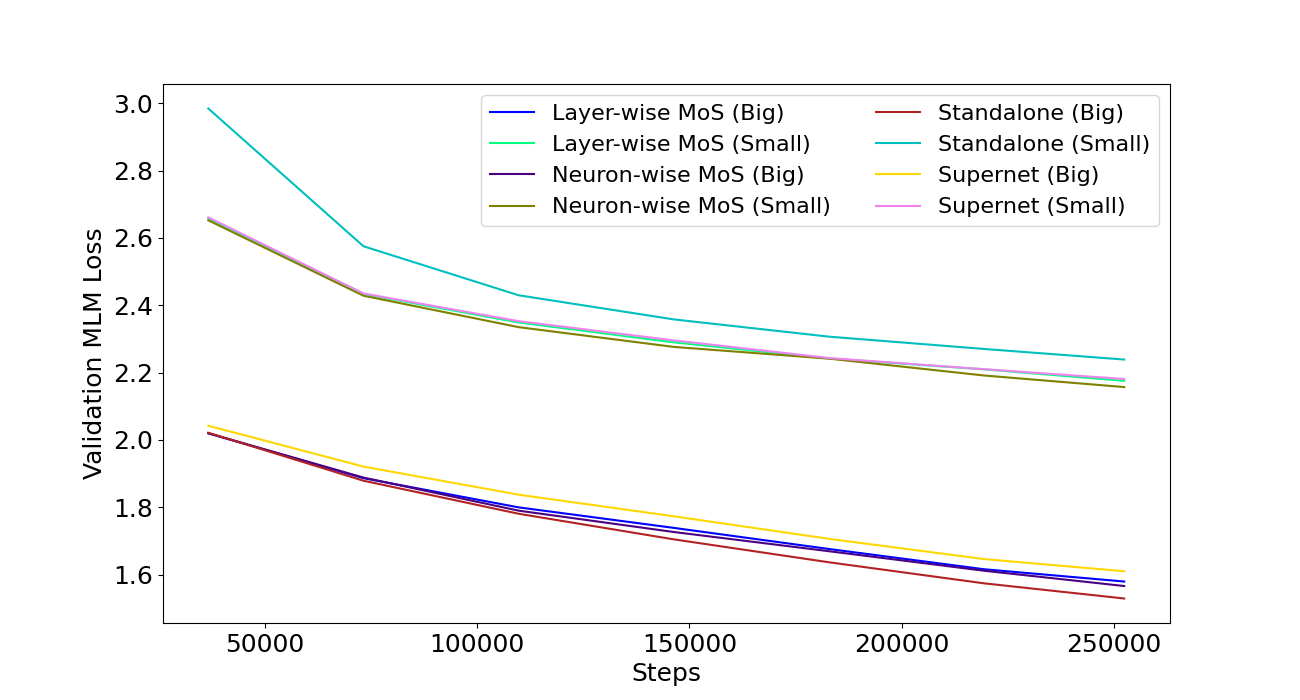}
\caption{Learning Curve - Training steps vs. Validation MLM loss. `Big' and `Small' correspond to the largest and the smallest BERT architecture respectively from the search space of SuperShaper. `Standalone' and `Supernet' correspond to training from scratch and sampling from the supernet respectively. All the supernets are trained with sandwich training.}
\label{fig:appd_bert_learncurve}
\end{figure*}

Figure~\ref{fig:appd_bert_learncurve} shows the training steps versus validation MLM loss (learning curve) for the standalone BERT model and different supernet based BERT variants. The standalone model and the supernet are compared for the biggest architecture (big) and the smallest architecture (small) from the search space of SuperShaper~\citep{supershaper}. For the biggest architecture, the standalone model performs the best. For the smallest architecture, the standalone model is outperformed by all the supernet variants. In both cases, the proposed supernets (especially neuron-wise MoS) perform much better than the standard supernet.

\subsubsection{Architecture comparison of Neuron-wise MoS vs. AutoDistil}
\label{sec:bert_arch_comparison}
Table~\ref{tab:bert_arch_analysis_autodistil} shows the comparison of the BERT architecture designed by our proposed neuron-wise MoS with AutoDistil.

\begin{table*}
\scriptsize
\begin{center}
\begin{tabular}{cccccc} \toprule
\textbf{Standalone / Supernet} & \textbf{Model Size}  & \textbf{\#Layers}  & \textbf{\#Hidden Size} & \textbf{\#FFN Expansion Ratio} & \textbf{\#Heads}   \\
\midrule
BERT & 109M & 12 & 768 & 4 & 12 \\
\midrule
AutoDistil (proxy) & 6.88M & 7 & 160 & 3.5 & 10 \\
Neuron-wise MoS  & 5M & 12 & 120 & 2.0 & 6 \\ \midrule
Neuron-wise MoS  & 10M & 12 & 180 & 3.5 & 6 \\ \midrule
AutoDistil (agnostic) & 26.8M & 11 & 352 & 4 & 10 \\
Neuron-wise MoS  & 26.8M & 12 & 372 & 2.5 & 6 \\ \midrule
Neuron-wise MoS  & 30M & 12 & 384 & 3 & 6 \\ \midrule
AutoDistil (proxy) & 50.1M & 12 & 544 & 3 & 9 \\
Neuron-wise MoS  & 50M & 12 & 504 & 3.5 & 12 \\ 
\bottomrule
\end{tabular}
\caption{Architecture comparison of the best architecture designed by the neuron-wise MoS with AutoDistil~\citep{autodistill} and BERT-Base~\citep{devlin-etal-2019-bert}.}
\label{tab:bert_arch_analysis_autodistil}
\end{center}
\end{table*}

\subsubsection{Fair comparison of Neuron-wise MoS w.r.t SoTA with MNLI}
\label{sec:bert_arch_comparison_incl_mnli}
\begin{table*}[t]
\scriptsize
\begin{center}
\begin{tabular}% {p{0.95in}p{0.4in}p{0.55in}p{0.25in}p{0.25in}p{0.25in}p{0.25in}p{0.25in}p{0.25in}p{0.25in}c} 
{ccccccccccc}
\toprule
\textbf{Supernet} & \textbf{\#Params}  & \textbf{\#Steps}  & \textbf{MNLI} & \textbf{CoLA} & \textbf{MRPC} & \textbf{SST2} & \textbf{QNLI}  & \textbf{QQP}  & \textbf{RTE} & \textbf{Avg. GLUE}     \\ \midrule
NAS-BERT & 5M & 125K & 74.4 & 19.8 & 79.6 & \textbf{87.3} & \textbf{84.9} & 85.8 & 66.7 & 71.2 \\
Neuron-wise MoS  & 5M & 0 & \textbf{75.5} & \textbf{28.3} & \textbf{82.7} & 86.9 & 84.1 & \textbf{88.5} & \textbf{68.1} & \textbf{73.4} \\ \midrule
NAS-BERT & 10M & 125K & 76.4 & 34.0 & 79.1 & \textbf{88.6} & \textbf{86.3} & 88.5 & \textbf{66.7} & 74.2 \\
Neuron-wise MoS & 10M & 0 & \textbf{77.2} & \textbf{34.7} & \textbf{81.0} & 88.1 & 85.1 & \textbf{89.1} & \textbf{66.7} & \textbf{74.6} \\ \midrule
AutoDistil (agnostic) & 26.8M & 0 & \textbf{82.8} & 47.1 & 87.3 & \textbf{90.6} & \textbf{89.9} & \textbf{90.8} & 69.0 & 79.6 \\
Neuron-wise MoS & 26.8M & 0 &  80.7 & \textbf{52.7} & \textbf{88.0} & 90.0 & {87.7} & 89.9 & \textbf{78.1} & \textbf{81.0} \\ \midrule
NAS-BERT & 30M & 125K & 81.0 & 48.7 & 84.6 & 90.5 & \textbf{88.4} & \textbf{90.2} & 71.8 & {79.3} \\
Neuron-wise MoS & 30M & 0 & \textbf{81.6} & \textbf{51.0} & \textbf{87.3} & \textbf{91.1} & 87.9 & \textbf{90.2} & \textbf{72.2} & \textbf{80.2} \\ \midrule
Neuron-wise MoS & 50M & 0 & 82.4 & {55.0} & 88.0 & {91.9} & 89.0 & 90.6 &  {75.4} & {81.8} \\ 
\bottomrule
\end{tabular}
\caption{Comparison of neuron-wise MoS with NAS-BERT and AutoDistil (agnostic) for different model sizes ($\leq50M$ parameters) based on GLUE validation performance. We include results on MNLI task. For fair comparison, we drop AutoDistil (proxy), which directly uses MNLI task for architecture selection. Neuron-wise MoS improves over the baselines in all model sizes, in terms of average GLUE. For MNLI task, neuron-wise MoS improves over the baselines in most model sizes.}
\label{tab:bert_comparison_to_sota_with_mnli}
\end{center}
\end{table*}
We compare neuron-wise MoS with NAS-BERT and AutoDistil (agnostic) for different model sizes ($\leq50M$ parameters) based on GLUE validation performance. In Table~\ref{tab:bert_comparison_to_sota_with_mnli}, we include results on MNLI task. For fair comparison, we drop AutoDistil (proxy), which directly uses MNLI task for architecture selection. Neuron-wise MoS improves over the baselines in all model sizes, in terms of average GLUE. For MNLI task, neuron-wise MoS improves over the baselines in most model sizes.

\subsubsection{BERT results with different random seeds}
\label{sec:bert_diff_seeds}

\begin{table*}
\scriptsize
\begin{center}
\begin{tabular}{cccccc} \toprule
\textbf{Seeds} & \multicolumn{2}{c}{\textbf{Seed 1}} & \multicolumn{2}{c}{\textbf{Seed 2}} \\ 
\textbf{Model} & \textbf{CoLA} & \textbf{RTE} & \textbf{CoLA} & \textbf{RTE} & \textbf{Average} \\ \midrule
Standalone & \textbf{59.03} & 71.53 & \textbf{58.04} & 72.22 & 65.21 \\
Supernet (Sandwich) & 57.58 & 73.26 & 57.1 & 72.92 & 65.22 \\
Layer-wise MoS & 57.62 & \textbf{77.08} & 56.3 & \textbf{76.74} & \textbf{66.91} \\
\bottomrule
\end{tabular}
\caption{BERT results on CoLA and RTE with different random seeds. Layer-wise MoS improves over baselines in RTE and degrades over baselines in CoLA consistently across both seeds.}
\label{tab:bert_diff_seeds}
\end{center}
\end{table*}

\update{Table~\ref{tab:bert_diff_seeds} displays BERT results on CoLA and RTE with various random seeds. Layer-wise MoS consistently enhances performance over baselines in RTE and diminishes performance compared to baselines in CoLA for both seeds. The BERT architecture (67M parameters) corresponds to the top model from the pareto front of Supernet (Sandwich) in SuperShaper's search space (consistent with Table~\ref{tab:bert_supernetvsscratch}).}

\subsection{Additional Experiments - Efficient Machine Translation}
\label{sec:appendix_experiments_mt}

\subsubsection{Machine translation benchmark data}
\label{sec:mt_bench_data}
Table~\ref{tab:dataset_statistics} shows the statistics of three machine translation datasets: WMT'14 En-De, WMT'14 En-Fr, and WMT'19 En-De.

\begin{table*}
\scriptsize
\begin{center}
\begin{tabular}{ccccccc} \toprule
\textbf{Dataset} & \textbf{Year} & \textbf{Source Lang} & \textbf{Target Lang} & \textbf{\#Train} & \textbf{\#Valid} & \textbf{\#Test} \\ \midrule
WMT & 2014 & English (en) & German (de) & 4.5M & 3000 & 3000 \\
WMT & 2014 & English (en) & French (fr) & 35M & 26000 & 26000 \\
WMT & 2019 & English (en) & German (de) & 43M & 2900 & 2900 \\
\bottomrule
\end{tabular}
\caption{Machine translation benchmark data.}
\label{tab:dataset_statistics}
\end{center}
\end{table*}

\subsubsection{Training settings and metrics}
\label{sec:mt_train_settings}
The training settings for both supernet and standalone models are the same: $40K$ training steps, Adam optimizer, a cosine learning rate scheduler, and a warmup of learning rate from $10^{-7}$ to $10^{-3}$ with cosine annealing. The best checkpoint is selected based on the validation loss, while the performance of the MT model is evaluated based on BLEU. The beam size is four with length penalty of 0.6. {The architecture encoding $a$ is a list of following 10 values:
\begin{enumerate}
    \item \textit{Encoder embedding dimension} corresponds to embedding dimension of the encoder.
    \item \textit{Encoder \#layers} corresponds to number of encoder layers.
    \item \textit{Average encoder FFN. intermediate dimension} corresponds to average of FFN intermediate dimension across encoder layers.
    \item \textit{Average encoder self attention heads} corresponds to average of number of self attention heads across encoder layers.
    \item \textit{Decoder embedding dimension} corresponds to embedding dimension of the decoder.
    \item \textit{Decoder \#Layers} corresponds to number of decoder layers.
    \item \textit{Average Decoder FFN. Intermediate Dimension} corresponds to average of FFN intermediate dimension across decoder layers.
    \item \textit{Average decoder self attention heads} corresponds to average of number of self attention heads across decoder layers.
    \item \textit{Average decoder cross attention heads} corresponds to average of number of cross attention heads across decoder layers.
    \item \textit{Average arbitrary encoder decoder attention} corresponds to average number of encoder layers attended by cross-attention heads in each decoder layer (-1 means only attend to the last layer, 1 means attend to the last two layers, 2 means attend to the last three layers).
\end{enumerate}}

\begin{comment}
\subsubsection{Comparison of ranking quality between supernet and standalone}
\label{sec:mt_kendall}
Table~\ref{tab:mt_kendall} shows the kendall rank correlation coefficient between the supernet and the standalone model BLEU performance for 15 random architectures from the MT search space. The ranking of architectures obtained by the proposed supernets seem largely similar to that obtained by the baseline supernets.

\begin{table*}
\scriptsize
\begin{center}
\begin{tabular}{p{0.8in}p{0.7in}p{0.7in}p{0.7in}} \toprule
\textbf{Dataset} & {\textbf{WMT'14 En-De}} & {\textbf{WMT'14 En-Fr}} & {\textbf{WMT'19 En-De}} \\
\textbf{Supernet} &  \textbf{Kendall} & \textbf{Kendall} &  \textbf{Kendall} 
\\ \midrule
HAT  & 0.81 & 0.63 & 0.71 \\
Supernet (Sandwich) & 0.81 & 0.63 & 0.87 \\
Arch. Experts & 0.54 & 0.73 & 0.87 \\
Neuron Experts & 0.71 &  0.85 &  0.81 \\
\bottomrule
\end{tabular}
\caption{Kendall rank correlation coefficient between the supernet and the standalone model BLEU performance for 15 random architectures from the MT search space.}
\label{tab:mt_kendall}
\end{center}
\end{table*}
\end{comment}

\subsubsection{Supernet vs. Standalone performance plot}
\label{sec:mt_supernet_vs_standalone}
\begin{figure*}[t!]
    \centering
    \begin{subfigure}[t]{0.4\textwidth}
        \centering
        \includegraphics[height=1.5in, width=2.6in]{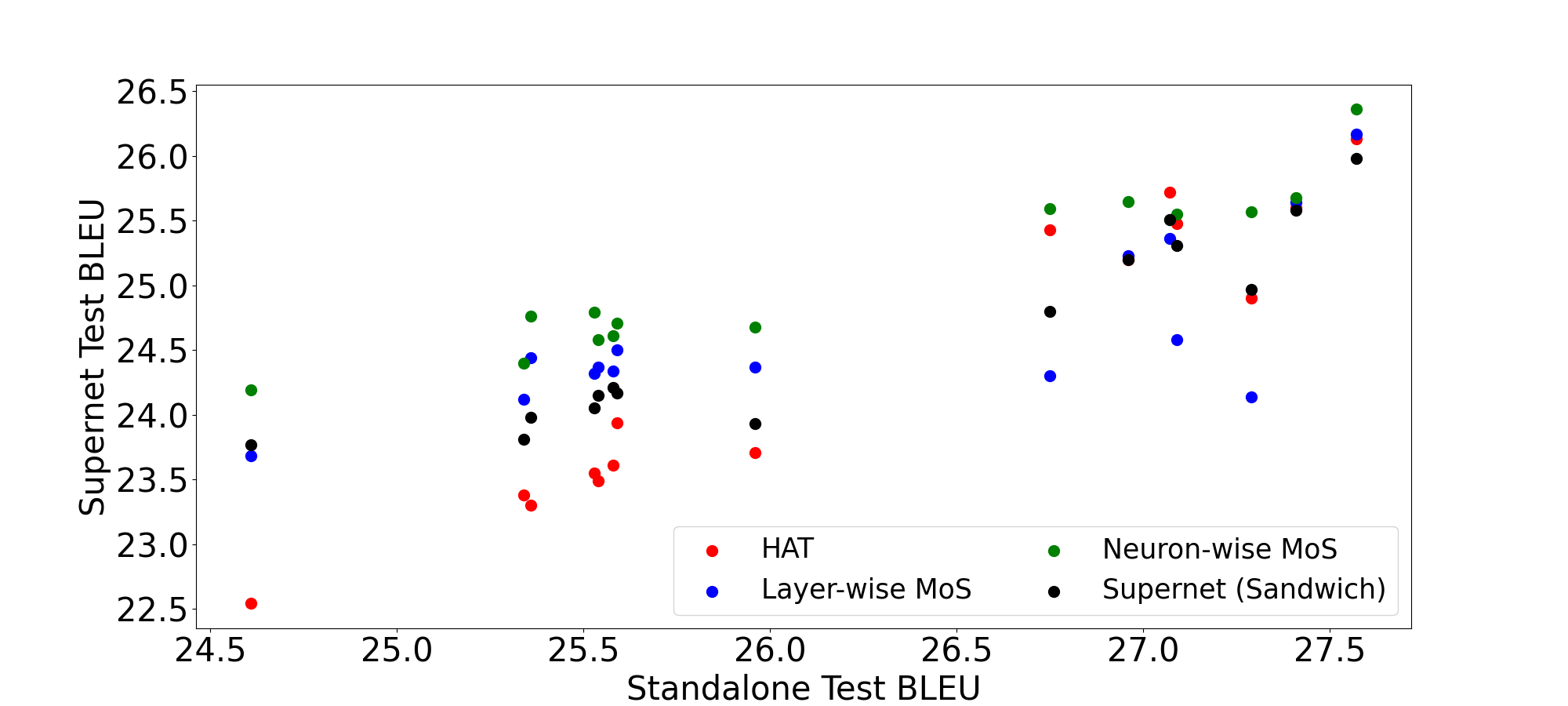}
        \caption{WMT'14 En-De}
    \end{subfigure}
    ~ 
    \begin{subfigure}[t]{0.4\textwidth}
        \centering
        \includegraphics[height=1.5in, width=2.6in]{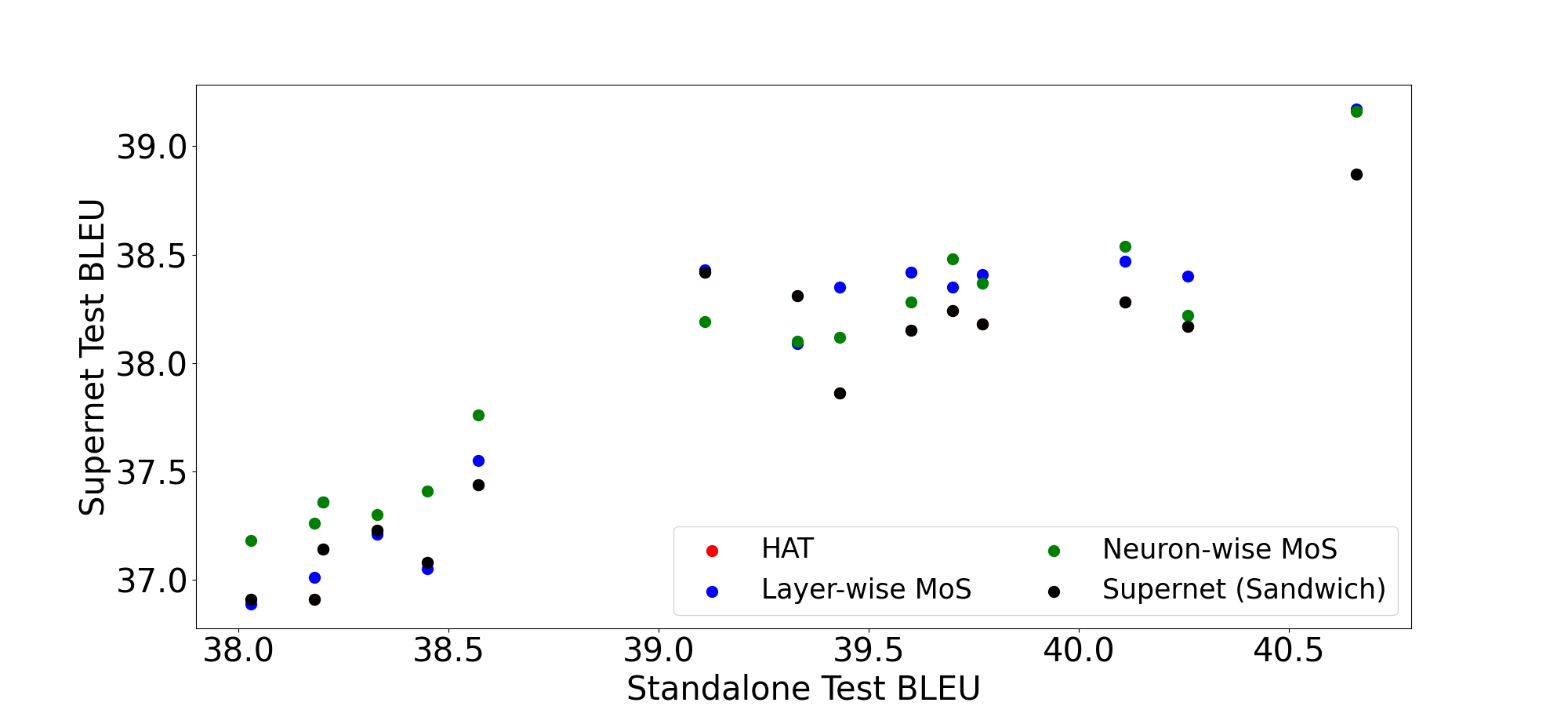}
        \caption{WMT'14 En-Fr}
    \end{subfigure}
    ~ 
    \begin{subfigure}[t]{1.0\textwidth}
        \centering
        \includegraphics[height=1.6in, width=2.6in]{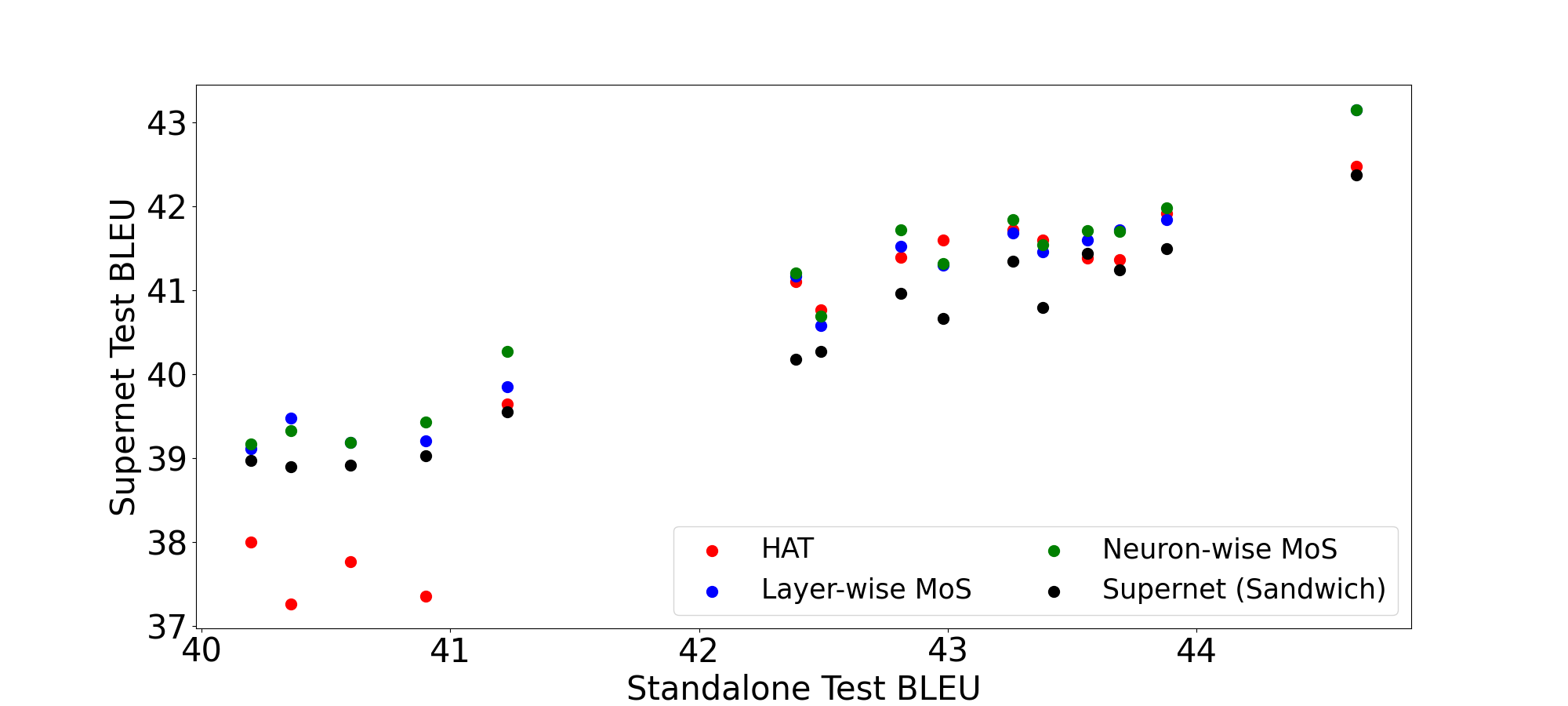}
        \caption{WMT'19 En-De}
    \end{subfigure}
    \caption{Supernet vs. Standalone model performance for 15 random architectures from MT search space. Supernet performance is obtained by evaluating the architecture-specific weights extracted from the supernet. Standalone model performance is obtained by training the architecture from scratch to convergence and evaluating it.}
    \label{fig:mt_supernetvsscratch}
\end{figure*}

Figure~\ref{fig:mt_supernetvsscratch} displays the supernet vs. the standalone performance for 15 randomly sampled architectures on all the three tasks. Neuron-wise MoS excel for almost all the top performing architectures ($\geq26.5$  and $\geq42.5$ standalone BLEU for WMT'14 En-De and WMT'19 En-De respectively), which indicates that the models especially in the pareto front can benefit immensely from neuron level specialization.  

% \hcnotes{why the x y labels seems to be streched? better to put a diagonal dashed line. I feel the table is easier to understand with some explanation, we can put this figure into appendix.}

\subsubsection{HAT Settings}
\label{sec:hat_settings}

\noindent\textbf{\underline{Evolutionary search}}: The settings for the evolutionary search algorithm include: 30 iterations, population size of 125, parents population of 25, crossover population of 50, and mutation population of 50 with 0.3 mutation probability. 

\noindent\textbf{\underline{Latency estimator}}: The latency estimator is developed in two stages. First, the latency dataset is constructed by measuring the latency of 2000 randomly sampled architectures directly on the user-defined hardware (NVIDIA V100 GPU). Latency is the time taken to translate a source sentence to a target sentence (source and target sentence lengths of 30 tokens each). For each architecture, 300 latency measurements are taken, outliers (top 10\% and bottom 10\%) are removed, and the rest (80\%) is averaged. Second, the latency estimator is a 3 layer multi-layer neural network based regressor, which is trained using encoding and latency of the architecture as features and labels respectively.

\subsubsection{Additional training steps to close the gap vs. performance}
\label{sec:addtrain_2data}
\begin{figure*}[t!]
    \centering
    \begin{subfigure}[t]{0.3\textwidth}
        \centering
        \includegraphics[height=1.3in, width=2.15in]{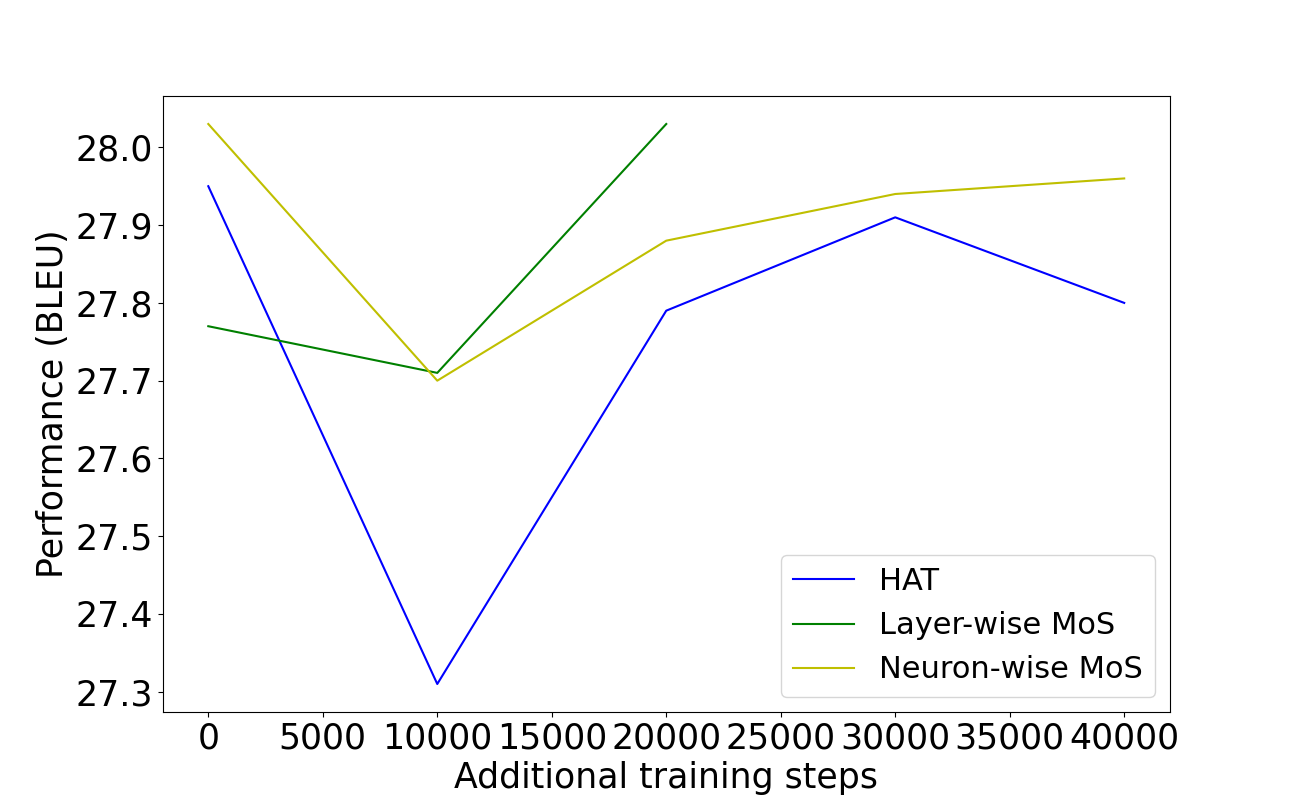}
        \caption{100ms}
    \end{subfigure}
    ~ 
    \begin{subfigure}[t]{0.3\textwidth}
        \centering
        \includegraphics[height=1.3in, width=2.15in]{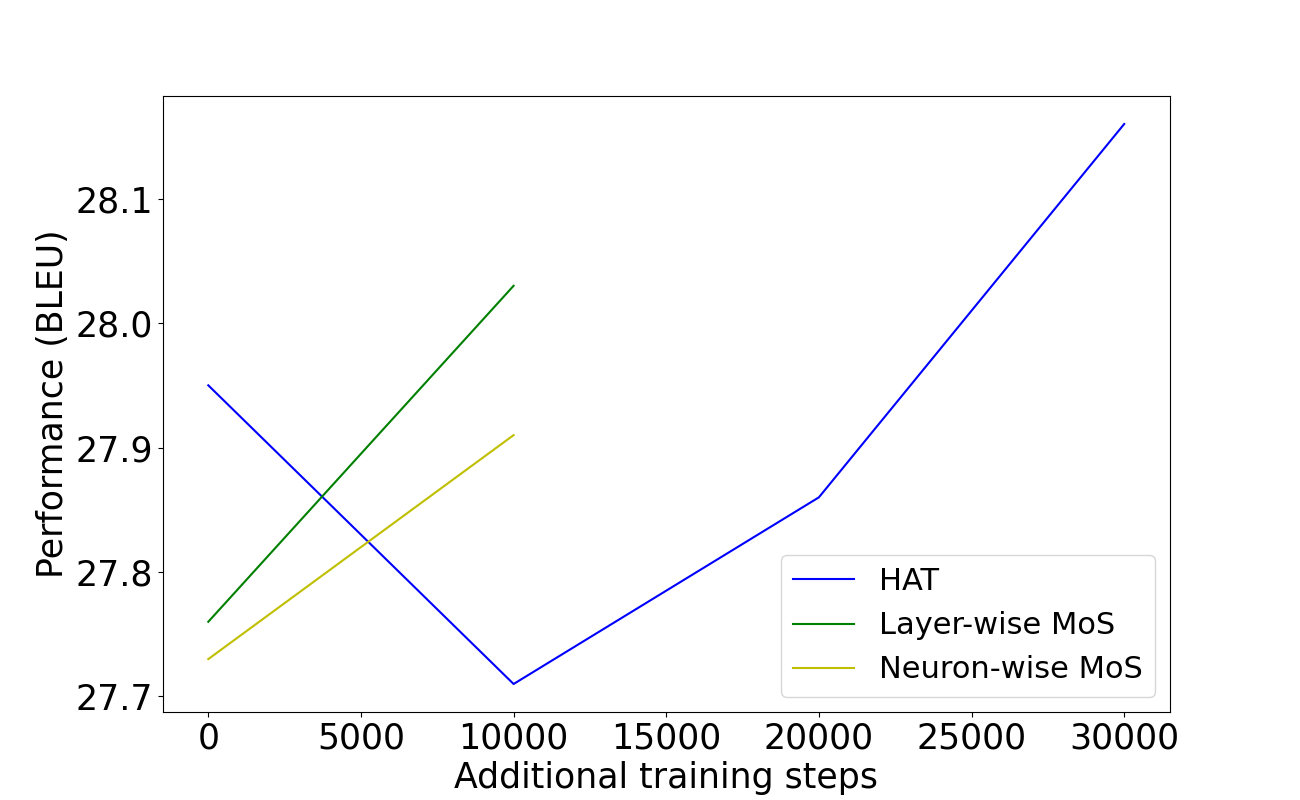}
        \caption{150ms}
    \end{subfigure}
    ~ 
    \begin{subfigure}[t]{0.3\textwidth}
        \centering
        \includegraphics[height=1.3in, width=2.15in]{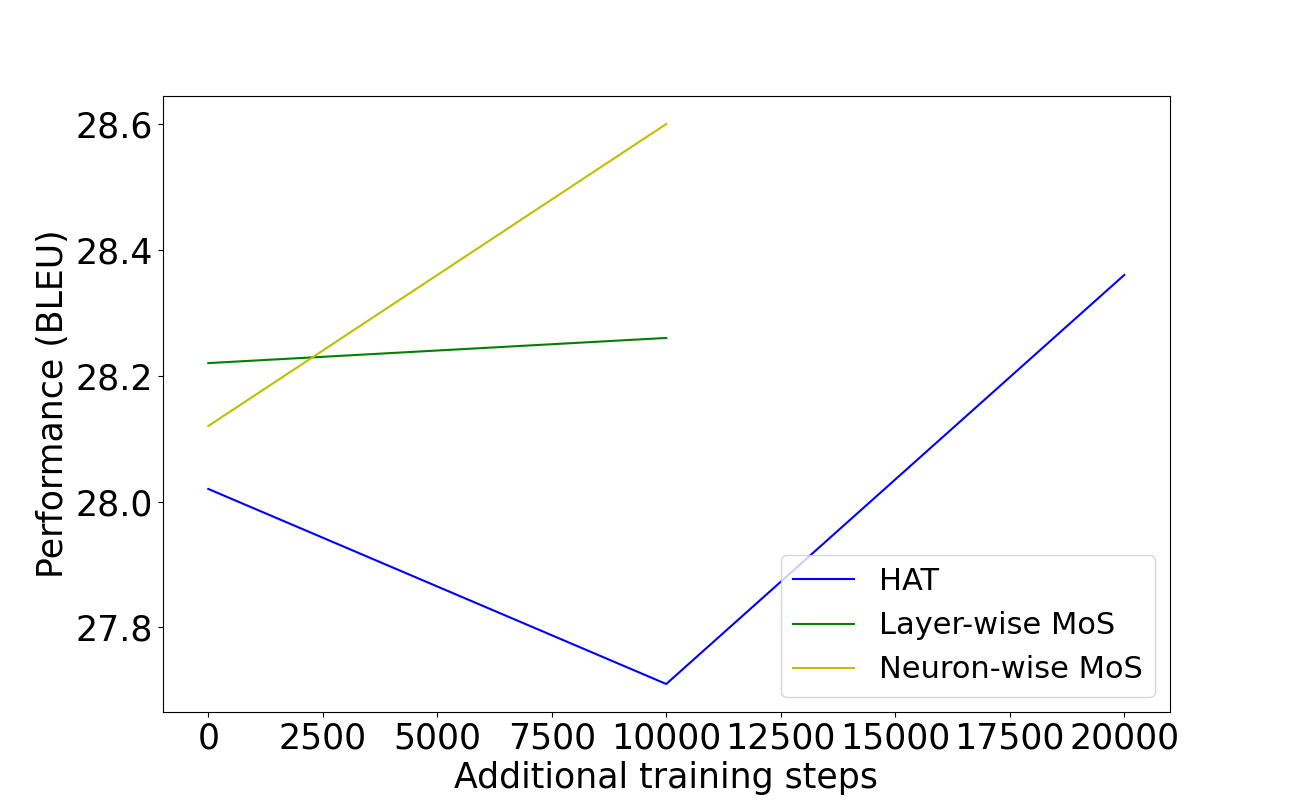}
        \caption{200ms}
    \end{subfigure}
    \caption{Additional training steps to close the supernet - standalone gap vs. performance for different latency constraints on the WMT'14 En-De dataset.}
    \label{fig:mt_addtrainsteps_wmt14ende}
\end{figure*}

\begin{figure*}[t!]
    \centering
    \begin{subfigure}[t]{0.3\textwidth}
        \centering
        \includegraphics[height=1.3in, width=2.15in]{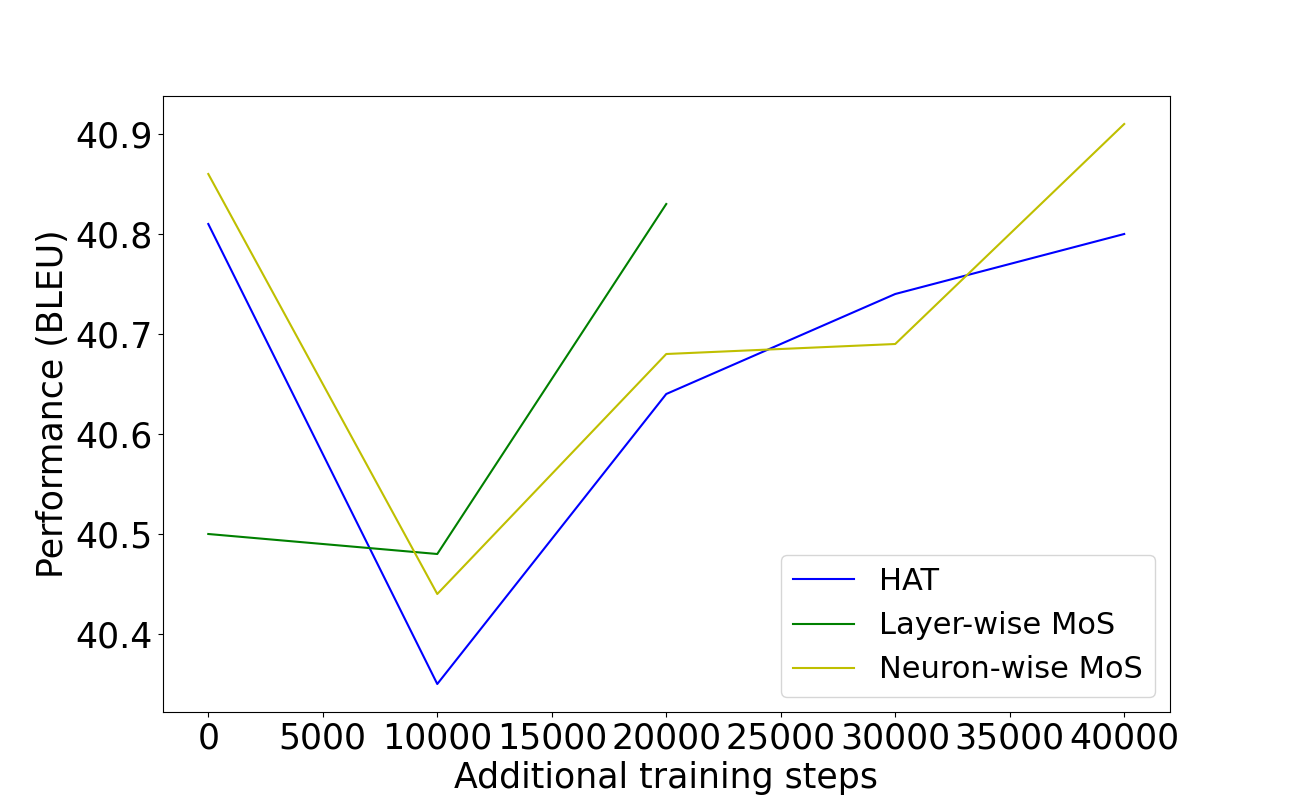}
        \caption{100ms}
    \end{subfigure}
    ~ 
    \begin{subfigure}[t]{0.3\textwidth}
        \centering
        \includegraphics[height=1.3in, width=2.15in]{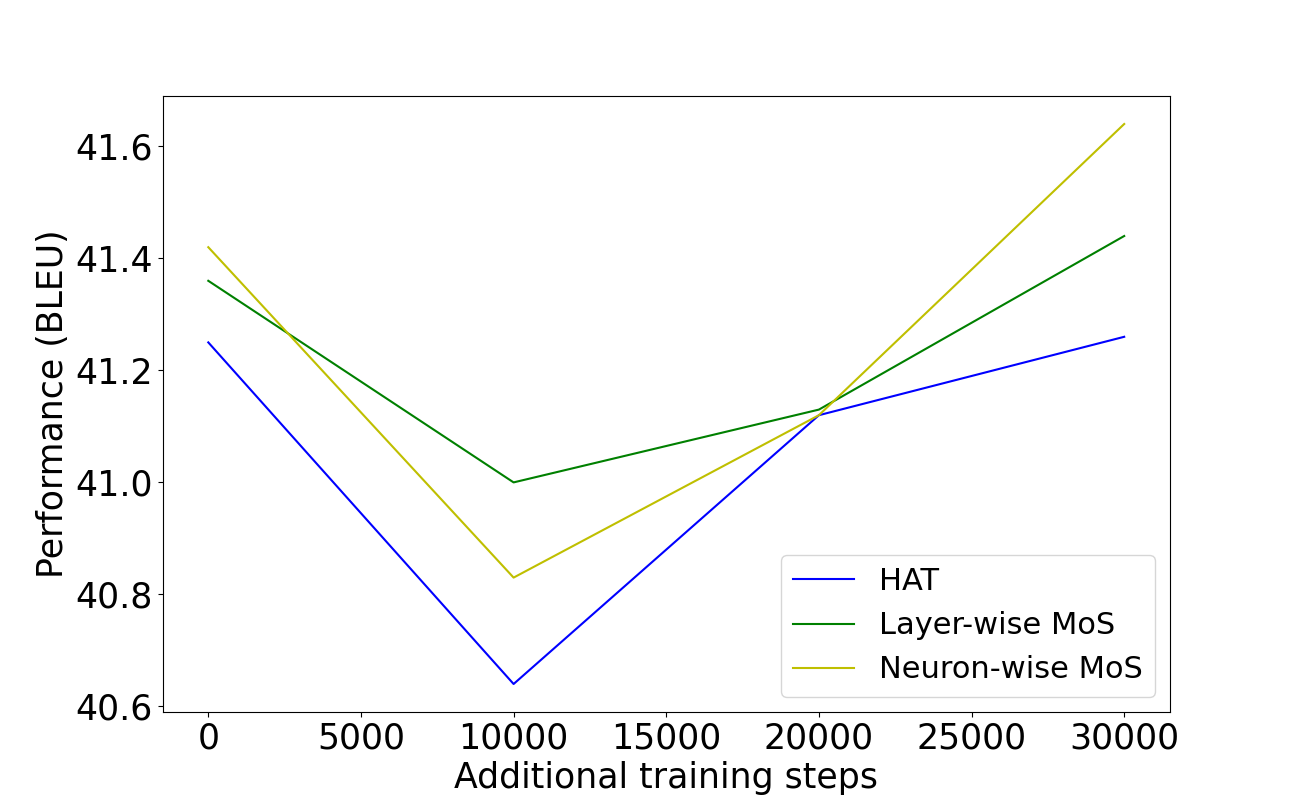}
        \caption{150ms}
    \end{subfigure}
    ~ 
    \begin{subfigure}[t]{0.3\textwidth}
        \centering
        \includegraphics[height=1.3in, width=2.15in]{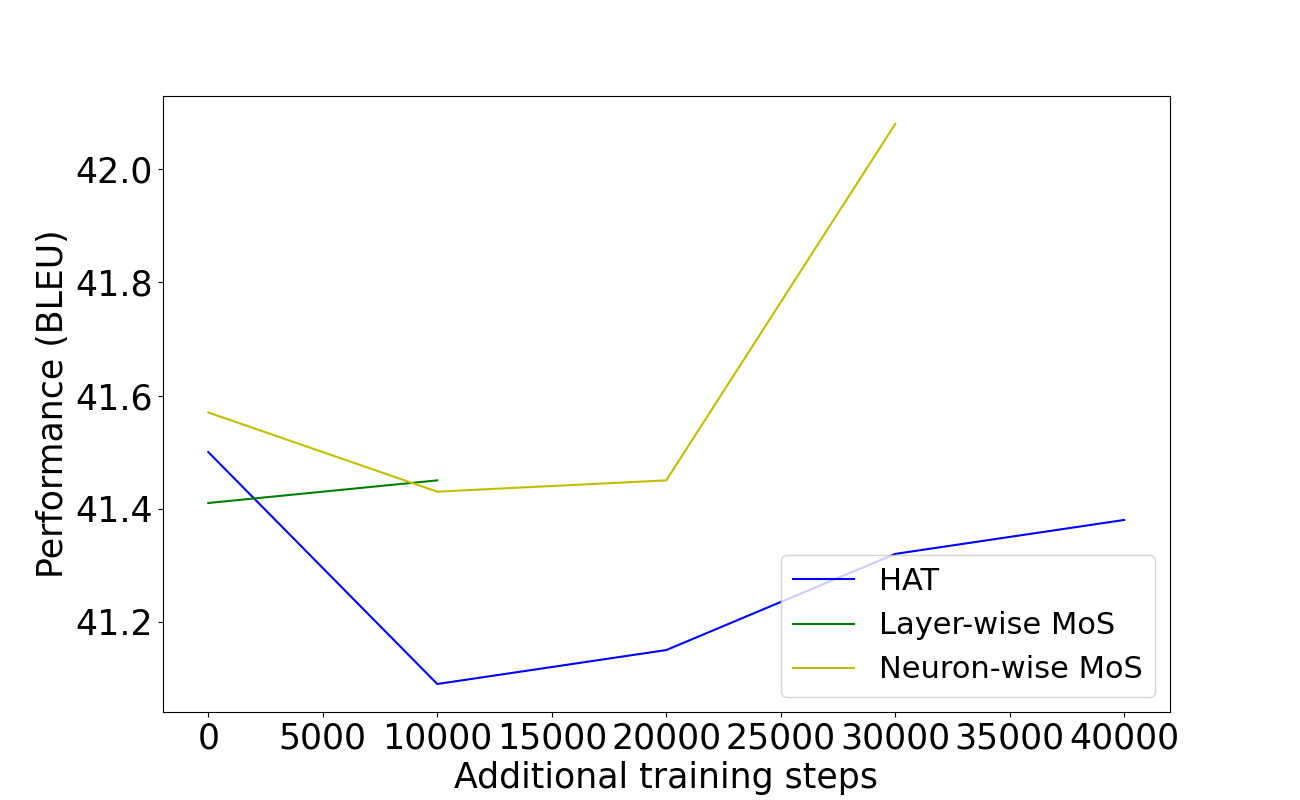}
        \caption{200ms}
    \end{subfigure}
    \caption{Additional training steps to close the supernet - standalone gap vs. performance for different latency constraints on the WMT'14 En-Fr dataset.}
    \label{fig:mt_addtrainsteps_wmt14enfr}
\end{figure*}

\begin{figure*}[t!]
    \centering
    \begin{subfigure}[t]{0.3\textwidth}
        \centering
        \includegraphics[height=1.3in, width=2.15in]{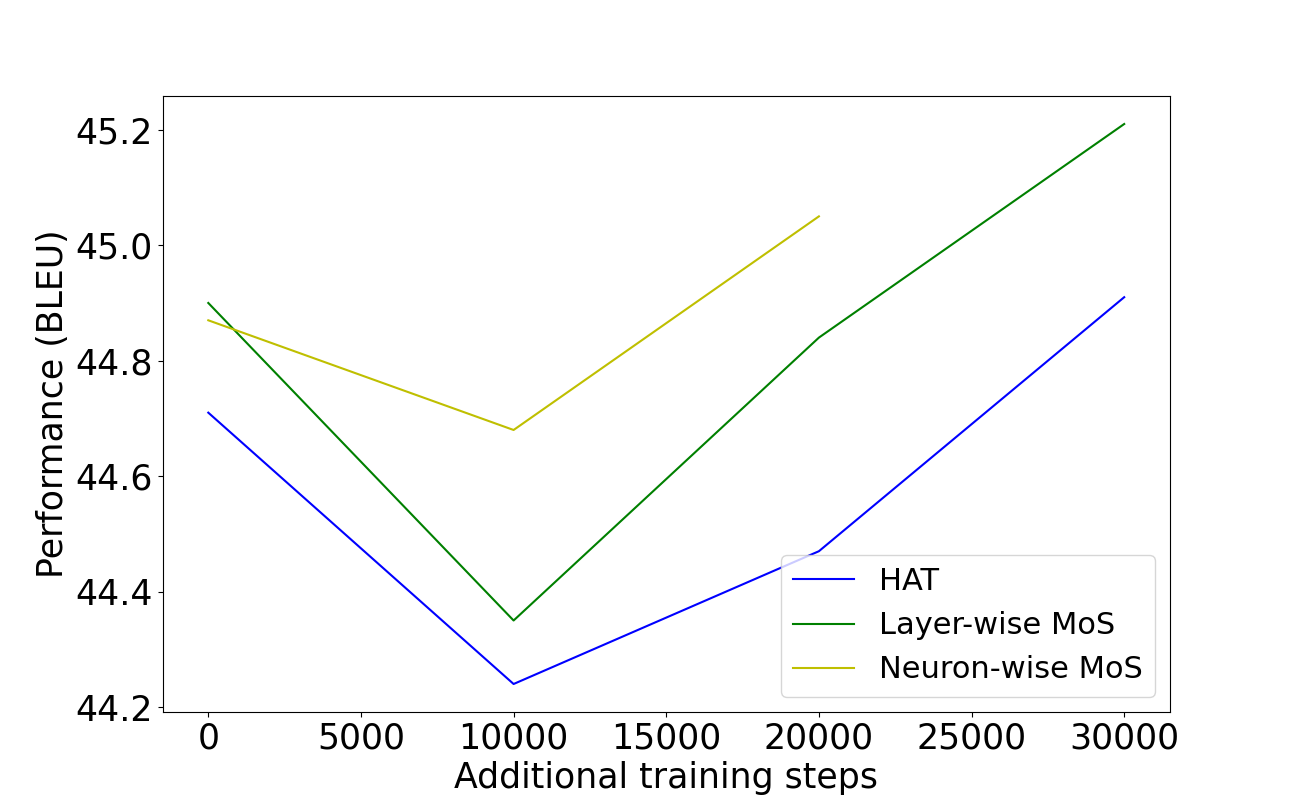}
        \caption{100 ms}
    \end{subfigure}
    ~ 
    \begin{subfigure}[t]{0.3\textwidth}
        \centering
        \includegraphics[height=1.3in, width=2.15in]{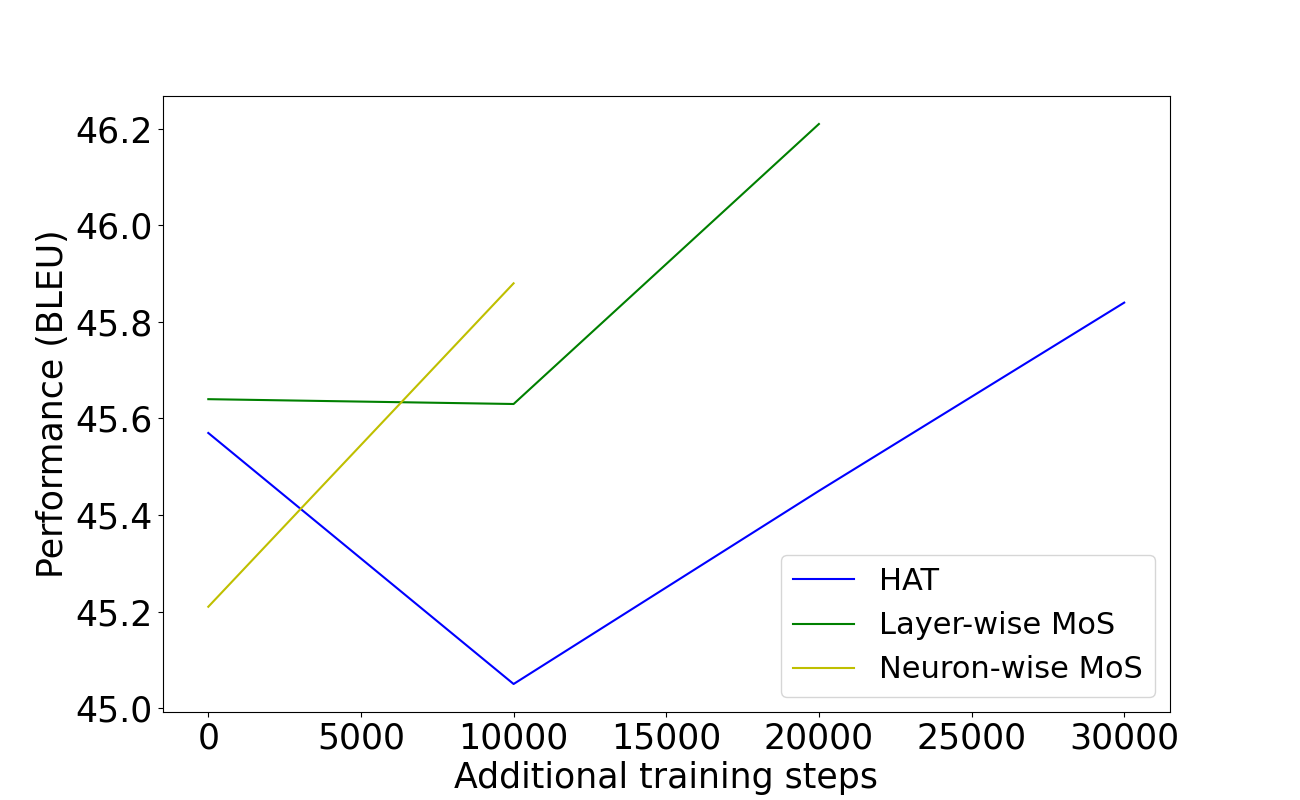}
        \caption{150 ms}
    \end{subfigure}
    ~ 
    \begin{subfigure}[t]{0.3\textwidth}
        \centering
        \includegraphics[height=1.3in, width=2.15in]{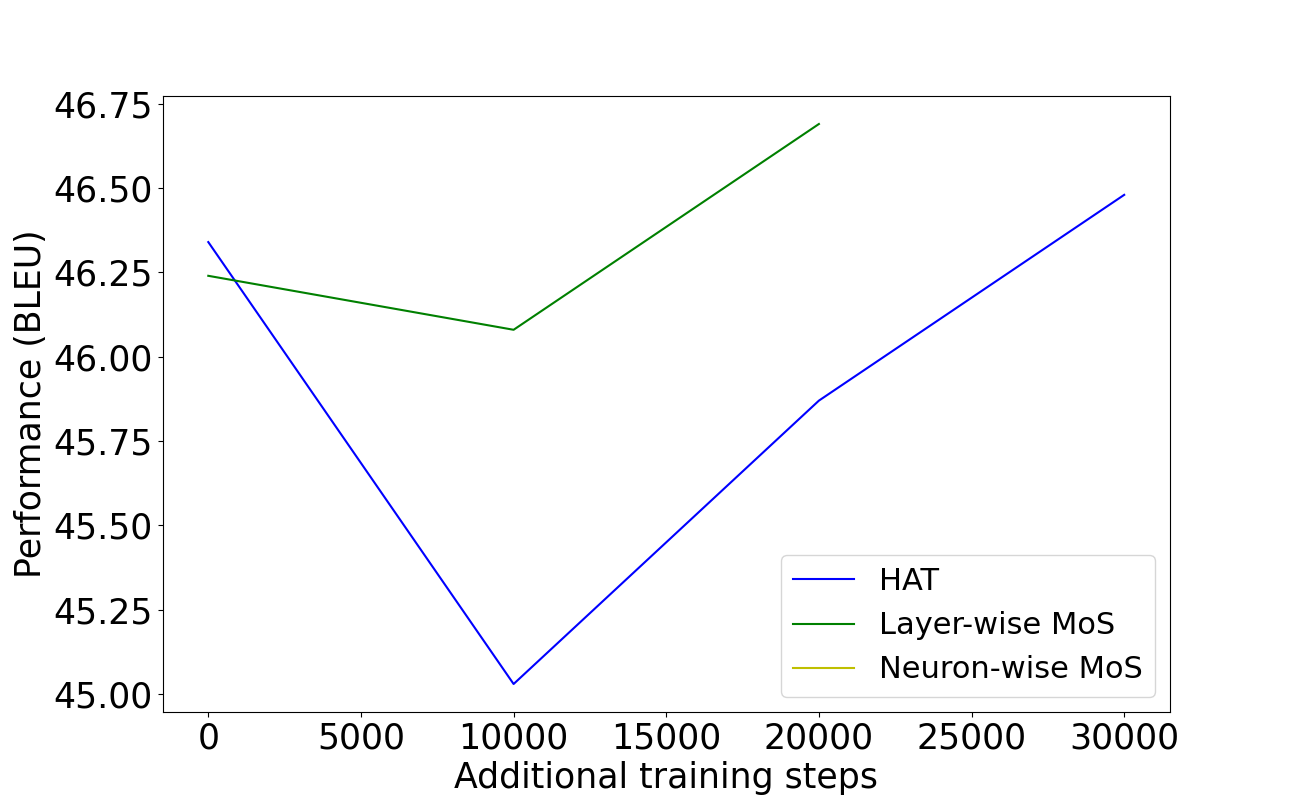}
        \caption{200 ms}
    \end{subfigure}
    \caption{Additional training steps to close the supernet - the standalone gap vs. performance for different latency constraints on the WMT'19 En-De dataset. For $200$ ms latency constraint, neuron-wise MoS closes the gap without additional training.}
    \label{fig:mt_addtrainsteps}
\end{figure*}

Figure~\ref{fig:mt_addtrainsteps_wmt14ende},  Figure~\ref{fig:mt_addtrainsteps_wmt14enfr}, and Figure~\ref{fig:mt_addtrainsteps} show the additional training steps vs. BLEU for different latency constraints on the WMT'14 En-De task, WMT'14 En-Fr and WMT'19 En-De tasks respectively.

% \hcnotes{show a horizontal dashed line of the standalone BLEU, maybe only HAT standalone BLEU to avoid multiple overlapping lines in the figure} \hcnotes{explain why there is no yellow line in figure (c)}

\subsubsection{Evolutionary Search - Stability}
\label{sec:evosearch_mt_stability}
We study the initialization effects on the stability of the pareto front outputted by the evolutionary search for different supernets. Table~\ref{tab:appd_mt_evosearch_stability} displays sampled (direct) BLEU and latency of the models in the pareto front for different seeds on the WMT'14 En-Fr task. The differences in the latency and BLEU across seeds are mostly marginal. This result highlights that the pareto front outputted by the evolutionary search is largely stable for all the supernet variants.

\begin{table*}
\scriptsize
\begin{center}
\begin{tabular}{cccccccc} \toprule
\textbf{Supernet / Pareto Front} &  &  \multicolumn{2}{c}{\textbf{Model 1}}  & \multicolumn{2}{c}{\textbf{Model 2}} & \multicolumn{2}{c}{\textbf{Model 3}} \\ 
 & \textbf{Seed} & \textbf{Latency} & \textbf{BLEU} & \textbf{Latency} & \textbf{BLEU} & \textbf{Latency} & \textbf{BLEU}   \\ 
\midrule
HAT (SPOS) & 1 & 96.39 & 38.94 & 176.44 & 39.26 & 187.53 & 39.16 \\
HAT (SPOS) & 2 & 98.91 & 38.96 & 159.87 & 39.20 & 192.11 & 39.09 \\
HAT (SPOS) & 3 & 100.15 & 38.96 & 158.67 & 39.24 & 189.53 & 39.16 \\
% \midrule
% HAT (Sandwich) & 1 & 98.47 & 39.06 & 159.19 & 39.96 & 226.75 & 40.96 \\
% HAT (Sandwich) & 2 & 115.40 & 39.14 & 159.87 & 40.18 & 189.27 & 40.88 \\
% HAT (Sandwich) & 3 & 118.31 & 39.29 & 158.37 & 40.03 & 233.89 & 40.94 \\
 \midrule
Layer-wise MoS & 1 & 99.42 & 39.34 & 158.68 & 40.29 & 205.55 & 41.24 \\
Layer-wise MoS & 2 & 99.60 & 39.32 & 156.48 & 40.29 & 209.80 & 41.13 \\
Layer-wise MoS & 3 & 119.65 & 39.32 & 163.17 & 40.36 & 208.52 & 41.18 \\
 \midrule
Neuron-wise MoS & 1 & 97.63 & 39.55 & 200.17 & 40.02 & 184.09 & 41.04 \\
Neuron-wise MoS & 2 & 100.46 & 39.55 & 155.96 & 40.04 & 188.87 & 41.15 \\
Neuron-wise MoS & 3 & 100.47 & 39.57 & 157.26 & 40.04 & 190.40 & 41.17 \\
\bottomrule
\end{tabular}
\caption{Stability of the evolutionary search w.r.t. different seeds on the WMT'14 En-Fr task. Search quality is measured in terms of latency and sampled (direct) supernet performance (BLEU) of the models in the pareto front.}
\label{tab:appd_mt_evosearch_stability}
\end{center}
\end{table*}

\begin{table}
\scriptsize
\begin{center}
\begin{tabular}{cc} \toprule
\textbf{\# layers in router function} & \textbf{BLEU ($\uparrow$)} \\ \midrule
2-layer & \textbf{26.61} \\
3-layer & 26.14 \\
4-layer & 26.12 \\
\bottomrule
\end{tabular}
\caption{Validation BLEU of different router functions for neuron-wise MoS on the WMT'14 En-De task. }
\label{tab:mt_diffrouterfn}
\end{center}
\end{table}
% \hcnotes{we can put this to appendix and only mentioned that we found 2 layer is the best, if the space is limited.}

% impact of different router fn
\subsubsection{Impact of different router function}
\label{sec:diffroutfn}
Table~\ref{tab:mt_diffrouterfn} displays the impact of varying the number of hidden layers in the router function for neuron-wise MoS on the WMT'14 En-De task. Two hidden layers provide the right amount of router capacity, while adding more hidden layers results in steady performance drop.

\subsubsection{Impact of increasing the number of expert weights `m'}
\label{sec:mt_vary_m}
{Table~\ref{tab:mt_vary_m} displays the impact of increasing the number of expert weights `m' for the WMT'14 En-Fr task, where the architecture for all the supernets is the top architecture from the pareto front of HAT for the latency constraint of $200$ ms. Under the standard training budget ($40K$ steps for MT), the performance of layer-wise MoS does not seem to improve by increasing `m' from 2 to 4. Increasing `m' introduces too many parameters, which might necessitate a significant increase in the training budget (e.g., 2 times more training steps than the standard training budget). For fair comparison with existing literature, we use the standard training budget for all the experiments. We will investigate the full potential of the proposed supernets by combining larger training budget (e.g., $\geq 200K$ steps) and larger number of expert weights (e.g., $\geq 16$ expert weights) in future work.}

\begin{table}
\scriptsize
\begin{center}
\begin{tabular}{cccc} \toprule
\textbf{Supernet} & \textbf{m} & \textbf{BLEU ($\uparrow$)} & \textbf{Supernet GPU Memory ($\downarrow$)}  \\ \midrule
HAT & - & 39.13 &  11.4 GB \\
Layer-wise MoS & 2 & \textbf{40.55} & 15.9 GB \\
Layer-wise MoS & 4 & 40.33 & 16.1 GB  \\
\bottomrule
\end{tabular}
\caption{{Impact of increasing the number of expert weights `m' for the WMT'14 En-Fr task. The architecture is the top model from the pareto front of HAT for the latency constraint of $200$ ms.}}
\label{tab:mt_vary_m}
\end{center}
\end{table}

\subsubsection{SacreBLEU vs. BLEU}
\label{sec:sacrebleu}
{We use the standard BLEU~\citep{papineni-etal-2002-bleu} to quantify the performance of supernet following HAT for a fair comparison. In Table~\ref{tab:sacrebleu}, we also experiment with SacreBLEU~\citep{sacrebleu}, where the similar trend of MoS yielding better performance for a given latency constraint holds true.} 
% \laksCom{Well, on quite similar: on SACREBLEU, Layerwise does better than Neuronwise, like BLEU. Are we able to explain this?}\gan{Not understanding the question. On both BLEU  and sacrebleu, neuron-wise is better than layer-wise MoS.}

\begin{table}
\scriptsize
\begin{center}
\begin{tabular}{ccc} \toprule
\textbf{Supernet} & \textbf{BLEU ($\uparrow$)} & \textbf{SacreBLEU ($\uparrow$)} \\ \midrule
HAT & 26.25 & 25.68 \\
Layer-wise MoS & 27.31 & 26.7 \\
Neuron-wise MoS & \textbf{27.59} & \textbf{27.0} \\
\bottomrule
\end{tabular}
\caption{{Performance of supernet as measured by BLEU and SacreBLEU for the latency constraint of $150$ ms on the WMT'14 En-De task.}}
\label{tab:sacrebleu}
\end{center}
\end{table}

\subsubsection{Breakdown of the overall time savings}
\label{sec:mt_overall_time_breakdown}
{Table~\ref{tab:mt_overall_time_breakdown} shows the breakdown of the overall time savings of MoS supernets versus HAT for computing pareto front for the WMT'14 En-De task. The latency constraints include $100$ ms, $150$ ms, $200$ ms. MoS have an overall GPU hours savings of at least 20\% w.r.t. HAT, thanks to significant savings in additional training time (45\%-51\%).}

\begin{table*}
\scriptsize
\begin{center}
\begin{tabular}{cc|ccc} \toprule
\textbf{Supernet} & \textbf{Overall Time ($\downarrow$)} & \textbf{Supernet Training Time ($\downarrow$)} & \textbf{Search Time ($\downarrow$)} & \textbf{Additional Training Time ($\downarrow$)} \\ \midrule
HAT & 508 hours & \textbf{248 hours} & \textbf{3.7 hours} & 256 hours \\ 
Layer-wise MoS & 407 hours (20\%) &  262 hours (-5.6\%) & 4.5 hours (-21.6\%) & 140 hours (45.3\%) \\
Neuron-wise MoS & \textbf{394 hours (22\%)} & 266 hours (-7.3\%) & 4.3 hours (-16.2\%) & \textbf{124 hours (51.6\%)} \\
\bottomrule  
\end{tabular}
\caption{{Breakdown of the overall time savings of MoS supernets vs. HAT for computing pareto front (latency constraints: $100$ ms, $150$ ms, $200$ ms) for the WMT'14 En-De task. Overall time (measured as single NVIDIA V100 hours) includes supernet training time, search time, and additional training time for the optimal architectures.} {Savings in parentheses.}}
\label{tab:mt_overall_time_breakdown} 
\end{center}
\end{table*}
% \laksCom{It may be helpful to show the savings on each component of the breakdown.}\gan{Added.} 

\subsubsection{Codebase}
\label{sec:code}
{We share the codebase at: \url{https://github.com/UBC-NLP/MoS}, which can be used to reproduce all the results in this paper. For both BERT and machine translation evaluation benchmarks, we add a README file that contains the following instructions: (i) environment setup (e.g., software dependencies), (ii) data download, (iii) supernet training, (iv) search,  and (v) subnet retraining.}